\documentclass{article}
\usepackage{ijcai26}

\usepackage{times}
\usepackage{soul}
\usepackage{url}
\usepackage[hidelinks]{hyperref}
\usepackage[utf8]{inputenc}
\usepackage[small]{caption}
\usepackage{graphicx}
\usepackage{amsmath}
\usepackage{amsthm}
\usepackage{booktabs}
\usepackage{algorithm}
\usepackage{algorithmic}
\usepackage[switch]{lineno}
\usepackage{mathtools}
\usepackage{amssymb}
\usepackage{natbib}
\usepackage{enumitem}
\usepackage{multirow}
\usepackage{array} 
\usepackage{tabularx} 
\usepackage{ragged2e}
\usepackage{microtype}
\usepackage{subcaption}
\usepackage{tikz}
\usetikzlibrary{arrows.meta,positioning,calc,trees,decorations.pathreplacing,fit,shapes.geometric,arrows.meta}
\tikzset{
  feat/.style={rectangle,draw,minimum width=4.5mm,minimum height=4.5mm,rounded corners=0pt},
  seen/.style={feat,fill=black!15},
  unseen/.style={feat,fill=white},
  op/.style={rectangle,draw,rounded corners=0pt,align=center,inner sep=2pt},
  decision/.style={circle,draw,inner sep=1.2pt},
  leaf/.style={ellipse,draw,inner sep=1.2pt}
}

\definecolor{mydarkblue}{rgb}{0,0.08,0.45}
\hypersetup{ %
pdftitle={},
pdfsubject={},
pdfkeywords={},
pdfborder=0 0 0,
pdfpagemode=UseNone,
colorlinks=true,
linkcolor=mydarkblue,
citecolor=mydarkblue,
filecolor=mydarkblue,
urlcolor=mydarkblue,
}

\newtheorem{proposition}{Proposition}

\title{A Survey on Active Feature Acquisition Strategies}

\author{
Linus Aronsson$^1$\And
Arman Rahbar$^1$\And
Morteza Haghir Chehreghani$^1$\\
\affiliations
$^1$Chalmers University of Technology and University of Gothenburg\\
\emails
\{linaro, armanr, morteza.chehreghani\}@chalmers.se
}

\begin{document}

\maketitle

\begin{abstract}
Active feature acquisition (AFA) studies how to sequentially acquire features for each data instance to trade off predictive performance against acquisition cost. This survey offers the first unified treatment of AFA via an explicit partially observable Markov decision process (POMDP) formulation. We place this formulation in the broader literature on optimal information acquisition and, more specifically, in a family of structured POMDPs (for example, information-gathering and sensing POMDPs) whose assumptions and algorithmic tools directly apply to AFA. This connection provides a common language for comparing problem settings and methods, and it highlights where AFA can leverage established results in structured POMDP planning and approximation. Building on this perspective, we present an up-to-date taxonomy of AFA methods that (roughly) mirrors standard approaches to solving POMDPs: (i) embedded cost-aware predictors (notably cost-sensitive decision trees and ensembles), (ii) model-based methods that plan using learned probabilistic components, (iii) model-free methods that learn acquisition policies from simulated episodes, and (iv) hybrid methods that combine the strengths of model-based and model-free approaches. We argue that this POMDP-centric view clarifies connections among existing methods and motivates more principled algorithm design. Since much prior work is heuristic and lacks formal guarantees, we also outline routes to guarantees by connecting AFA to adaptive stochastic optimization. We conclude by highlighting open challenges and promising directions for future research.

\end{abstract} 

\section{Introduction}
\label{sec:introduction}

Many predictive systems rely on features that are costly, slow, invasive, or privacy sensitive to obtain. In medical decision support, ordering all tests for every patient can be expensive and burdensome, and it may delay treatment \citep{GorryBarnett1968, 10.5555/1622826.1622838}. In personalized marketing and recommender systems, collecting all behavioral signals can be costly and raises privacy concerns \citep{Jeckmans2013}. In interactive troubleshooting, diagnostic queries (for example, checks, measurements, or questions) consume time and user effort \citep{10.1145/203330.203341}. Similar constraints arise in sensor systems and robotics, where measurements require power, time, or motion \citep{stachniss2005information,10.5555/1641503.1641516, 9899480}. These settings motivate \emph{active feature acquisition} (AFA), where an agent sequentially decides which feature values to acquire for each instance, and when to stop and predict, trading accuracy against feature acquisition cost. AFA is also known by other terms, including dynamic feature selection and instance-wise feature acquisition.

AFA differs fundamentally from standard (static) feature selection, which commits to a single fixed subset used for all instances \citep{DBLP:journals/jmlr/GuyonE03}. The advantage of AFA is most apparent when the informative features vary across instances. For example, suppose there is one feature that indicates which of many otherwise independent features is the only one relevant for predicting the label on the current instance. In order to be accurate on all instances, a static subset rule must include the indicator feature and all other features. In contrast, AFA learns an \emph{adaptive} acquisition policy: it can first acquire the indicator feature and then, based on its realized value, choose the next feature accordingly, achieving the same accuracy while using far fewer features per instance \citep{DBLP:conf/icml/ValanciusLO24}. See Figure~\ref{fig:afa-acquisition-costs} for an illustration of the adaptive acquisition procedure in AFA.

AFA is also distinct from several other feature selection paradigms.
(i) \emph{Online feature selection}, where features arrive over time and the learner maintains a compact set of informative features under memory or test-time constraints \citep{Hu2018OnlineFeatureSelection}.
(ii) \emph{Train-time active feature acquisition} \citep{1565772, 10.1145/1089827.1089828, saar2009active}, where the objective is to actively acquire features on the training set to trade off acquisition cost against predictive performance on a test set, typically assuming that all features are available at test time.
A related line of work instead actively acquires \emph{labels} for each instance (usually referred to simply as active learning) \citep{settles2009active, 10.1145/3472291}, or both features and labels \citep{saar2009active}.
(iii) \emph{Static instance-wise feature acquisition}, where features are selected per instance at test time even though all features are available, often to promote sparsity or interpretability \citep{pmlr-v80-chen18j, yoon2018invase}.

\begin{figure*}[t]
\centering
\begin{tikzpicture}[
    >=Stealth,
    every node/.style={font=\small},
    node distance=0cm
]

\def\barsep{1.85cm}        
\def\predsep{2.20cm}       
\def\rowsep{1.10cm}        
\def\dotsgap{0.30cm}       
\def\afterdotsgap{0.40cm}  

\tikzset{
  arrlabel/.style={
    midway,
    font=\small,
    inner sep=1.2pt,
    fill=white,
    fill opacity=0.85,
    text opacity=1
  }
}

\def\cellsize{0.30cm}      
\def\barw{1.00}

\newcommand{\featurebar}[4]{%
  \node[inner sep=0pt, outer sep=0pt, #1] (#2) {%
    \begin{tikzpicture}[x=\cellsize,y=\cellsize]
      \foreach \k/\shade in {#4}{
        \path[fill=gray!\shade] (0,\k) rectangle (\barw,\k+1);
      }
      \foreach \k in {0,...,5}{
        \draw[line width=0.40pt, rounded corners=0pt] (0,\k) rectangle (\barw,\k+1);
      }
    \end{tikzpicture}
  };
  \node[below=0.12cm of #2, font=\scriptsize] {#3};
}

\newcommand{\featurebarL}[4]{%
  \node[inner sep=0pt, outer sep=0pt, #1] (#2) {%
    \begin{tikzpicture}[x=\cellsize,y=\cellsize]
      \foreach \k/\lbl in {0/$x_1$,1/$x_2$,2/$x_3$,3/$x_4$,4/$x_5$,5/$x_6$}{
        \node[anchor=east, font=\scriptsize] at (-0.32,\k+0.5) {\lbl};
      }
      \foreach \k/\shade in {#4}{
        \path[fill=gray!\shade] (0,\k) rectangle (\barw,\k+1);
      }
      \foreach \k in {0,...,5}{
        \draw[line width=0.40pt, rounded corners=0pt] (0,\k) rectangle (\barw,\k+1);
      }
    \end{tikzpicture}
  };
  \node[below=0.12cm of #2, font=\scriptsize] {#3};
}

\featurebarL{at={(0,0)}}{i1bar0}{$S=\{1,2\}$}{0/25,1/95}
\featurebar{right=\barsep of i1bar0}{i1bar1}{$S=\{1,2,5\}$}{0/25,1/95,4/45}

\node[left=0.22cm of i1bar0, align=right] {Instance 1};

\node[right=\predsep of i1bar1,
      draw,
      inner xsep=5pt, inner ysep=3pt,
      align=left] (i1pred)
      {$\hat{y} = f(x_1,x_2,x_5)$\\{\scriptsize Total cost: $c_5 + \ell(\hat y, y)$}};

\draw[->] (i1bar0.east) --
  node[arrlabel, above]{\(\pi(x_S)=5\)}
  node[arrlabel, below]{cost \(c_5\)}
  (i1bar1.west);
\draw[->] (i1bar1.east) --
  node[arrlabel, above]{\(\pi(x_S)=\texttt{STOP}\)}
  (i1pred.west);

\featurebarL{below=\rowsep of i1bar0}{i2bar0}{$S=\{1,2\}$}{0/75,1/35}
\featurebar{right=\barsep of i2bar0}{i2bar1}{$S=\{1,2,3\}$}{0/75,1/35,2/60}
\featurebar{right=\barsep of i2bar1}{i2bar2}{$S=\{1,2,3,5\}$}{0/75,1/35,2/60,4/25}
\featurebar{right=\barsep of i2bar2}{i2bar3}{$S=\{1,2,3,5,6\}$}{0/75,1/35,2/60,4/25,5/80}

\node[left=0.22cm of i2bar0, align=right] {Instance 2};

\node[right=\predsep of i2bar3,
      draw,
      inner xsep=5pt, inner ysep=3pt,
      align=left] (i2pred)
      {$\hat{y} = f(x_1,x_2,x_3,x_5,x_6)$\\{\scriptsize Total cost: $c_3+c_5+c_6+\ell(\hat y, y)$}};

\draw[->] (i2bar0.east) --
  node[arrlabel, above]{\(\pi(x_S)=3\)}
  node[arrlabel, below]{cost \(c_3\)}
  (i2bar1.west);
\draw[->] (i2bar1.east) --
  node[arrlabel, above]{\(\pi(x_S)=5\)}
  node[arrlabel, below]{cost \(c_5\)}
  (i2bar2.west);
\draw[->] (i2bar2.east) --
  node[arrlabel, above]{\(\pi(x_S)=6\)}
  node[arrlabel, below]{cost \(c_6\)}
  (i2bar3.west);
\draw[->] (i2bar3.east) --
  node[arrlabel, above]{\(\pi(x_S)=\texttt{STOP}\)}
  (i2pred.west);

\node[below=\dotsgap of i2bar0] (dotsRow) {};
\node[left=0.95cm of dotsRow, align=right] {$\vdots$};

\featurebarL{below=\afterdotsgap of dotsRow}{inbar0}{$S=\{1,2\}$}{0/30,1/65}
\featurebar{right=\barsep of inbar0}{inbar1}{$S=\{1,2,5\}$}{0/30,1/65,4/80}
\featurebar{right=\barsep of inbar1}{inbar2}{$S=\{1,2,4,5\}$}{0/30,1/65,4/80,3/50}

\node[left=0.22cm of inbar0, align=right] {Instance \(n\)};

\node[right=\predsep of inbar2,
      draw,
      inner xsep=5pt, inner ysep=3pt,
      align=left] (inpred)
      {$\hat{y} = f(x_1,x_2,x_4,x_5)$\\{\scriptsize Total cost: $c_5+c_4+\ell(\hat y, y)$}};

\draw[->] (inbar0.east) --
  node[arrlabel, above]{\(\pi(x_S)=5\)}
  node[arrlabel, below]{cost \(c_5\)}
  (inbar1.west);
\draw[->] (inbar1.east) --
  node[arrlabel, above]{\(\pi(x_S)=4\)}
  node[arrlabel, below]{cost \(c_4\)}
  (inbar2.west);
\draw[->] (inbar2.east) --
  node[arrlabel, above]{\(\pi(x_S)=\texttt{STOP}\)}
  (inpred.west);

\node[draw, inner sep=5pt, anchor=north east] (legend)
  at ($(current bounding box.north east)+(1.2cm,-0.10cm)$) {%
  \begin{tikzpicture}[x=\cellsize,y=\cellsize]
    \draw[fill=white, line width=0.45pt, rounded corners=0pt] (0,0) rectangle (1,1);
    \node[anchor=west, font=\scriptsize] at (1.35,0.5) {Unobserved feature (white)};
    \draw[fill=gray!55, line width=0.45pt, rounded corners=0pt] (0,-1.6) rectangle (1,-0.6);
    \node[anchor=west, font=\scriptsize] at (1.35,-1.1) {Observed feature (shade indicates outcome)};
  \end{tikzpicture}
};

\end{tikzpicture}
\caption{Illustration of the adaptive acquisition process for active feature acquisition across instances. The goal is to find a policy $\pi$ and predictor $f$ that maximizes the quality of predictions while minimizing the cost of features acquired across instances. In this example, features 1 and 2 are assumed to always be available initially without cost (e.g., demographics such as age group of medical patient).}
\label{fig:afa-acquisition-costs}
\end{figure*}

At a conceptual level, AFA is a form of \emph{optimal information acquisition}, which can be categorized into (i)  \emph{non-adaptive} and (ii) \emph{adaptive} approaches \citep{golovin2011adaptive, chen2017phd}. In the context of \emph{feature selection}, standard static feature selection \citep{DBLP:journals/jmlr/GuyonE03} represents the non-adaptive case, while AFA corresponds to the adaptive case. Adaptive optimal information acquisition problems are typically instances of adaptive stochastic optimization and are often computationally intractable, which explains the prevalence of approximations such as myopic value of information rules \citep{10.5555/1641503.1641516,golovin2011adaptive, chen2017phd}. AFA connects to a broad set of related fields that study cost aware information gathering, including sequential experimental design \citep{f1168103-f0ea-30c5-aa6c-babe695713d3, DBLP:journals/actanum/HuanJM24}, active learning \citep{10.1145/3472291}, multi-armed bandits \citep{slivkins2019introduction}, decision analysis \citep{4082064, 4082150}, information gathering in robotics and perception \citep{stachniss2005information,9899480} and operations research \citep{nemhauser1978analysis, frazier2009knowledge}.

As described in Section~\ref{sec:problem_formulation}, AFA can be formulated as a \emph{partially observable Markov decision process} (POMDP) \citep{ASTROM1965174}. Existing approaches largely differ in how they approximate the optimal action-value function in \eqref{eq:qstar_acquire}. Methods for decision-making in the broader POMDP literature are commonly organized by when computation is performed (offline versus online planning) and by whether the environment model is assumed known or must be learned (planning with a model versus reinforcement learning approaches, including model-based and model-free variants) \citep{10.5555/1622673.1622690, NIPS2010_edfbe1af}. While many AFA papers do not explicitly adopt POMDP terminology, their algorithmic choices map naturally onto these paradigms. The goal of this survey is to structure the AFA literature around the underlying POMDP formulation and the approximation decisions it entails. Accordingly, we group AFA methods into four categories (Table~\ref{tab:afa_methods_taxonomy}): (i) \emph{embedded cost-aware predictors}, such as cost-sensitive decision trees and ensembles; (ii) \emph{model-based methods} that learn or approximate the probabilistic components of the acquisition process and use them for model-based POMDP planning; (iii) \emph{model-free methods} that learn a policy or action-value function directly from simulated acquisition episodes; and (iv) \emph{hybrid methods} that combine complementary elements of model-based and model-free approaches.


There are two important historical precursors to AFA:
(i) decision-theoretic medical expert systems, with the Pathfinder system as a canonical example \citep{HeckermanHorvitzNathwani1992a}, and
(ii) cost-sensitive decision trees \citep{Nunez1991EG2,Norton1989}, which modify standard decision tree induction to account for feature-acquisition costs.
More generally, any decision tree induces a valid AFA policy because it performs instance-wise feature selection at inference time.
Decision trees are therefore a key related topic, and we discuss this connection in detail in Section~\ref{sec:embedded}.
Modern AFA generalizes these ideas to arbitrary predictors and richer policy classes, and it makes the underlying sequential decision structure explicit.
This, in turn, enables the use of generic planning and reinforcement learning (RL) methods.

Finally, \citep{attenberg2011selective} surveys selective data acquisition and briefly mentions AFA, and \citep{ZoisChelmis2024SDM} surveys feature selection and acquisition across multiple settings (including AFA to some extent). To the best of our knowledge, this is the first survey focused specifically on AFA that (i) formulates AFA in detail as an POMDP, (ii) uses this framework to unify the diverse literature under a common lens, and (iii) provides an up-to-date taxonomy of AFA methods.


\section{Problem Formulation}
\label{sec:problem_formulation}

In this section, we define the AFA problem. For each instance, an agent sequentially acquires feature values (incurring costs) and decides when to stop and predict.

\paragraph{Notation.}
Let $p(\mathbf{x},\mathbf{y})$ be the data distribution, where $\mathbf{x}=(\mathbf{x}_1,\dots,\mathbf{x}_d)\in\mathcal{X}$ and $\mathbf{y}\in\mathcal{Y}$. In this paper, we represent random variables using bold symbols (e.g. $\mathbf{x}, \mathbf{y}$), while their possible values are shown in regular font (i.e. $x, y \sim p(\mathbf{x}, \mathbf{y})$).
Acquiring feature $a\in[d]\triangleq\{1,\dots,d\}$ reveals $x_a$ at cost $c_a\in\mathbb{R}_+$.
For $S\subseteq[d]$, let $x_S=\{x_a:a\in S\} \in \mathcal{X}_S$, and $c(S)=\sum_{a\in S}c_a$.

\paragraph{Predictor under partial observability.}
A predictor must handle arbitrary subsets of observed features:
\begin{equation}
\label{eq:predictor_short}
f:\{(S,x_S)\mid S\subseteq[d],\,x_S\in\mathcal{X}_S\}\to\mathcal{Y}.
\end{equation}
In practice, $f$ is often \emph{probabilistic}, meaning it outputs a distribution $p(\mathbf{y}\mid x_S)$, from which a point prediction can be obtained (for example, $\hat y=\arg\max_y p(y\mid x_S)$). We write $f(x_S)$ as shorthand for $f(S, x_S)$.

In practice, \(f\) outputs \(p(\mathbf{y}\mid x_S)\) only if it is Bayes optimal. Throughout the survey, we continue to write \(p(\mathbf{y}\mid x_S)\), with the understanding that in practice we only have access to an approximation of this distribution. The same is true for any marginal/conditional probabilistic quantity based on the true data distribution $p(\mathbf{x},\mathbf{y})$ (e.g., $p(\mathbf{x}_a \mid x_S)$).

To apply $f$ to partially observed inputs, one typically maps $(S,x_S)$ to a fixed-dimensional representation. A common approach is \emph{masking}: represent the current information by a binary mask $m\in\{0,1\}^d$ together with a full-length feature vector $\tilde x\in\mathbb{R}^d$, where $m_a=1$ indicates that feature $a$ has been acquired and $m_a=0$ otherwise.
The input to $f$ is then a concatenation such as $(m,\tilde x)$.
The unobserved entries of $\tilde x$ can be filled with (i) a fixed placeholder value (for example, $0$) or (ii) \emph{imputed} values produced by an imputation rule or model, for example $\tilde x_a\sim p(\mathbf{x}_a\mid x_S)$. 

Another alternative is to use the fact that $f(x_S) = \mathbb{E}_{\mathbf{x}_U \mid x_S}[f(x_S,\mathbf{x}_U)]$, where $U = [d] \setminus S$ denotes the set of unobserved features. In practice, any estimate of $f(x_S)$ amounts to approximating this marginalization. If an imputation model is available, the expectation can be approximated via Monte Carlo sampling.

Finally, some methods use permutation-invariant deep set encoders, see discussion in Section \ref{sec:conclusion}.


\paragraph{Acquisition process.} The features to be acquired for an instance $x \in \mathcal{X}$ are determined \emph{adaptively} by a (stochastic) policy $\pi$. Let $(S,x_S)$ be the features acquired so far and $U = [d] \setminus S$ be remaining unobserved features. Let $B \geq 0$ be the feature acquisition budget that can be spent on each instance. The policy maps the current partial observation to an action: $\pi(S, x_S) \in \mathcal{A}_B(S)$, where

\begin{equation} \label{eq:actionspace}
    \mathcal{A}_B(S) \triangleq \{a\in U:\;c(S)+c_a\le B\} \cup \{\texttt{STOP}\}\},
\end{equation}

is the set of available actions. We also denote $\mathcal{F}_B(S) \triangleq \mathcal{A}_B(S) \setminus \{\texttt{STOP\}}$. In practice, the policy often outputs a probability distribution over available actions. If $\pi$ chooses $a\in U$, we pay the cost $c_a$ and observe $x_a$. Then, we update $S=S\cup\{a\}$ and $U=U\setminus\{a\}$. If $\pi$ chooses \texttt{STOP}, we output $\hat y=f(S,x_S)$ and terminate. Let $\pi[x] \subseteq [d]$ denote the final set of feature indices acquired when the policy stops on instance $x$. In practice, the policy $\pi$ is usually implemented via masking, similar to the predictor $f$. The acquisition process is visualized in Figure \ref{fig:afa-acquisition-costs}.

\paragraph{Optimization objective.} A standard objective trades off prediction loss and acquisition cost under a budget constraint:
\begin{equation}
\begin{aligned}
\label{eq:afa_obj_soft}
&\min_{f,\pi} 
\mathbb{E}_{\mathbf{x},\mathbf{y}} 
\mathbb{E}_{\pi} \left[
\ell \big(f(\mathbf{x}_{\pi[\mathbf{x}]}),\mathbf{y}\big)
+ \alpha c(\pi[\mathbf{x}])\right] \\
&\quad\text{s.t.}\quad
c(\pi[x])\le B\;\;\text{for all }x\in\mathcal{X}.
\end{aligned}
\end{equation}


where \( \alpha \ge 0 \) controls the cost--accuracy trade-off and \( \ell \) is a loss function. The term \( \alpha\, c(\pi[\mathbf{x}]) \) allows the acquisition cost to vary across instances, so more resources can be spent on instances that are harder to predict. In this sense, \( \alpha \) induces a \emph{soft} budget. By contrast, \( B \) imposes a \emph{hard} budget.

A soft budget is beneficial because it adapts the cost spent per instance, but choosing an appropriate value of \( \alpha \) can be unintuitive in many applications. A hard budget \( B \) is more natural when resources are strictly limited per instance and is often easier to interpret. Its drawback is that, for some instances, accurate prediction may require acquiring features whose total cost exceeds \( B \). Two common special cases in AFA are \( \alpha = 0 \) (only a hard budget) and \( B = c([d]) \) (all features are, in principle, acquirable for every instance, so only the soft-budget term matters).

Other possible formulations of AFA exist \citep{golovin2011adaptive, DBLP:journals/ml/JanischPL20}, but we omit them for brevity, since the formulation above is by far the most common (typically under one of the special cases).


%
\paragraph{Training.} 
Most AFA methods assume access to i.i.d.\ training samples $(x,y)\sim p(\mathbf{x},\mathbf{y})$ with fully observed features and labels (costs are only incurred at test time). The predictor $f$ and policy $\pi$ are learned on this fully observed data (either jointly or separately) and then deployed on test instances from the same distribution, where features are initially unobserved and must be acquired sequentially at a cost. The predictor $f$ is commonly handled in one of two ways: (i) it is pretrained on offline data by randomly sampling masks $S$, or (ii) it is trained jointly with the acquisition policy from on-policy experience, that is, using states encountered by the policy during learning.

Only a small fraction of work studies an \emph{online} variant in which the learner acquires features during training as well and updates $f$ and $\pi$ from streaming experience, so learning is cost-sensitive from the start rather than relying on fully observed training data (see e.g., \citep{kachuee2018opportunistic, ijcai2023p463, rahbar2025costefficient}).

\section{AFA-POMDP Formulation} \label{sec:pomdp_formulation}
In this section, we show that the optimization problem in \eqref{eq:afa_obj_soft} can be formulated as reward maximization in a corresponding episodic, finite-horizon POMDP. 

\paragraph{State and action space.}  Let $S \subseteq [d]$ be the set of observed features, and $U = [d] \setminus S$ the remaining set of unobserved features. The state space is defined as $\mathcal S = \{(S,x,y):S \in 2^{[d]},x\in\mathcal X,\ y\in\mathcal Y\}$. Note that we have $x = (x_U, x_S)$ for all \(S\) and \(U\). We can therefore write any state $(S,x,y)$ as $(S,x_S,x_U,y) \in \mathcal{S}$. This decomposition is useful because it separates the state into an observed component \((S,x_S)\) and an unobserved component \((y, x_U)\). We use 

\begin{equation} \label{eq:observedstate}
    s(x_S,y) \triangleq (S,x_S,x_U,y) \in \mathcal{S}
\end{equation}

to denote the state of instance $x$ with label $y$ when $x_S$ has been observed so far. For any state in $\mathcal{S}$ where the features in $S$ have been observed, the action space is $\mathcal{A}_B(S)$, as defined in \eqref{eq:actionspace}.


\paragraph{Transition and observation model.} Given a state $s(x_S,y) \in \mathcal S$, both the transition model and observation model are deterministic. If $a \in U$, the feature $x_a$ is deterministically observed from $x_U$. Subesequently, we transition deterministically to the state $(S\cup \{a\}, x_{S} \cup x_a, x_U \setminus x_a,y)$. If $a = \texttt{STOP}$, we transition to an absorbing terminal state and make a prediction of $\mathbf{y}$ given our observed evidence $(S,x_S)$.

\paragraph{Belief state and observation probability.} The belief state corresponds to our current belief of the latent part of the state $(x_U, y)$, given the observed information $(S,x_S)$, and is represented by the distribution $p(\mathbf{x}_U,\mathbf{y}\mid x_S)$. The belief state is hence represented by

\begin{equation} \label{eq:beliefstate}
    b(x_S) \triangleq (S, x_S, p(\mathbf{x}_U, \mathbf{y} \mid x_S)).
\end{equation}

The probability of observing $o \in \mathcal{X}_a$ after taking action $a \in U$ in the current belief state is:

\begin{align} 
    &p(\mathbf{x}_a = o \mid x_S) =  \mathbb{E}_{\mathbf{y} \mid x_S}[p(\mathbf{x}_a = o \mid x_S, \mathbf{y})] \label{eq:obs2} \\
    &= \mathbb{E}_{\mathbf{x}_{U \setminus \{a\}},\mathbf{y} \mid x_S}[p(\mathbf{x}_a = o \mid x_S,\mathbf{x}_{U \setminus \{a\}}, \mathbf{y})] \label{eq:obs1}
\end{align}






\paragraph{Belief update.} After taking action $a \in U$ and observing $o \in \mathcal{X}_a$ in current belief state $b(x_S)$, let 

\begin{equation} \label{eq:update}
    b(x_S,a,z) \triangleq (S \cup \{a\}, x_{S} \cup o, p(\mathbf{x}_{U \setminus \{a\}}, \mathbf{y} \mid x_S,\mathbf{x}_a = o))
\end{equation}
    
denote the updated belief state given the new evidence. The updated belief $p(\mathbf{x}_{U \setminus \{a\}}, \mathbf{y} \mid x_S,\mathbf{x}_a = o)$ can be obtained via Bayesian updating. For example, for the label $\mathbf{y}$, the update is

\begin{equation} \label{eq:bayes}
\begin{aligned}
    &p(\mathbf{y} \mid x_S,\mathbf{x}_a=o) = \frac{p(\mathbf{x}_a = o \mid x_S,\mathbf{y})p(\mathbf{y}\mid x_S)}{p(\mathbf{x}_a = o \mid x_S)}.
\end{aligned}
\end{equation}

\eqref{eq:bayes} is the standard belief update rule in POMDPs \citep{ASTROM1965174, 1307539f-051d-3d3c-a0d8-111443bed03f}.

\paragraph{Reward.} For any state $s(x_S,y) \in \mathcal{S}$,  assume each acquisition action $a\in U$ yields immediate reward $R(s(x_S,y),a) = -\alpha c_a$. In addition, the stopping action $a = \texttt{STOP}$ yields a terminal reward $R(s(x_S,y),\texttt{STOP}) = -\ell(f(x_S), y)$. For action $a = \texttt{STOP}$, the expected reward in current belief state $b(x_S)$ is

\begin{equation} \label{eq:reward}
\begin{aligned} 
    r(b(x_S), \texttt{STOP}) &=\mathbb{E}_{\mathbf{x}_U,\mathbf{y}\mid x_S} [R(s(x_S,\mathbf{y}),\texttt{STOP})] \\
    &= \mathbb{E}_{\mathbf{x}_U,\mathbf{y}\mid x_S}[-\ell(f(x_S, \mathbf{x}_U),  \mathbf{y})] \\
    &= \mathbb{E}_{\mathbf{y}\mid x_S}[-\ell(f(x_S), \mathbf{y})] .
\end{aligned}
\end{equation}

Similarly, for any action $a \in U$, we have $r(b(x_S), a) = R(s(x_S,y),a) = - \alpha c_a$ (since $R(s(x_S,y),a)$ does not depend on $x$ or $y$). In practice, the conditional distribution $p(\mathbf{y}\mid x_S)$ is unknown and must be approximated, typically by the predictive distribution returned by our model, i.e., $f(x_S) \approx p(\mathbf{y}\mid x_S)$. Under log loss this gives $r(b(x_S), \texttt{STOP}) = -H(\mathbf{y}\mid x_S) = \mathbb{E}_{\mathbf{y} \mid x_S}[\log p(\mathbf{y}\mid x_S)]$ (i.e., the entropy of $p(\mathbf{y}\mid x_S)$). Under $0$--$1$ loss we obtain $r(b(x_S), \texttt{STOP}) = \max_{y\in\mathcal{Y}} p(y\mid x_S)$. When $f$ is Bayes optimal, these quantities coincide with the negative conditional Bayes risk for the corresponding loss.

\paragraph{Optimal value function of belief-MDP.}  Any POMDP can be rewritten as a fully observable \emph{belief-MDP} whose state is the current belief. The optimal value function of the induced belief-MDP for the AFA-POMDP is shown below, and can be solved as a standard MDP. Specifically, we show the $k$-step truncated action-value function $V_k$. It computes the value of current belief state $b(x_S)$, where the $\texttt{STOP}$ action is forced after $k$ selections starting from $b(x_S)$. We can easily relax the truncation by considering $k=d$.  
\begin{equation}
\begin{aligned}
\label{eq:pomdp_bellman_concise}
&V_k(b(x_S))=\max\Big\{
\underbrace{V_0(b(x_S))}_{a=\texttt{STOP}}, \max_{a\in \mathcal{F}_B(S)}\Big\{ -\alpha c_a + \\
&\mathbb{E}_{\mathbf{x}_a\mid x_S}[V_{k-1}(b(x_S, a,\mathbf{x}_a))]\Big\}\Big\},
\end{aligned}
\end{equation}
where $V_0(b(x_S)) = r(b(x_S), \texttt{STOP})$, $p(\mathbf{x}_a\mid x_S)$ is given by \eqref{eq:obs2} and $b(x_S,a, \mathbf{x}_a)$ is the updated belief state as defined in \eqref{eq:update}. The updated belief given the new evidence is obtained via the Bayesian update discussed in \eqref{eq:bayes}.

\paragraph{Equivalence to the AFA objective.} Consider an acquisition episode for instance $(x,y) \sim p(\mathbf{x}, \mathbf{y})$. Let $T$ be the first step where $\texttt{STOP}$ is chosen. The undiscounted return for one acquisition episode is then $G_{\pi}(x,y)=\sum_{t=1}^{T}R(s_t,a_t) = \sum_{t=1}^{T-1}(-\alpha c_{a_t}) + R(s_T,\texttt{STOP})$, where $s_t = (S_t,x_{S_t},x_{U_{t}}, y)$ is the state at step $t$, and $a_t$ is the action taken at step $t$ by policy $\pi$. Moreover, we have that $-\mathbb{E}_{p(\mathbf{x},\mathbf{y})}
\mathbb{E}_{\pi}[G_{\pi}(\mathbf{x},\mathbf{y})]$ is exactly the AFA optimization objective in \eqref{eq:afa_obj_soft}. Hence, maximizing expected return across episodes is equivalent to minimizing \eqref{eq:afa_obj_soft}. This connection was made for the first time in \citep{dulac2011datum}, where they use model-free reinforcement learning to jointly learn the policy $\pi$ and predictor $f$.

\paragraph{Model-based vs.\ model-free methods for the AFA-POMDP.}
Any approach that seeks a policy $\pi$ for the AFA-POMDP must be able to compute the reward $R(s(x_S,y),\texttt{STOP}) = \ell(f(x_S), y)$ in \eqref{eq:reward}. Moreover, $R(s(x_S,y),\texttt{STOP})$ requires a prediction $\hat y = f(x_S)$ such that the loss $\ell(f(x_S),y)$ can be computed. This means that the predictive distribution $f(x_S) \approx p(\mathbf{y} \mid x_S)$ is always required, regardless of the method is model-based or model-free. When $f$ is not probabilistic, it may not return $p(\mathbf{y} \mid x_S)$ explicitly, but it must still be trained on data from the underlying distribution in order to produce a prediction $\hat y = f(x_S)$.

We therefore use the term \emph{model-based} to denote methods that explicitly model $p(\mathbf{x}_U \mid x_S)$, since this conditional distribution effectively determines the transition dynamics in the AFA-POMDP (see, e.g., \eqref{eq:pomdp_bellman_concise}). Conversely, methods we refer to as \emph{model-free} do not model $p(\mathbf{x}_U \mid x_S)$ explicitly, but they still require a predictor for $\mathbf{y}$ (e.g., access to $p(\mathbf{y} \mid x_S)$) in order to evaluate the stopping reward, $R(s(x_S,y),\texttt{STOP})$.

\paragraph{Quantities required for model-based planning.} The Bellman recursion in \eqref{eq:pomdp_bellman_concise} depends on two  distributions at each belief state $b(x_S)$. The first is the predictive distribution $p(\mathbf{y}\mid x_S)$, which determines the stopping value through
$r(b(x_S), \texttt{STOP})=\mathbb{E}_{\mathbf{y}\mid x_S} \left[-\ell \left(f(x_S),\mathbf{y}\right)\right]$
as in \eqref{eq:reward}. The second is, for every unacquired feature $a \in \mathcal{F}_B(S)$, the observation distribution $p(\mathbf{x}_a\mid x_S)$ used to compute $\mathbb{E}_{\mathbf{x}_a\mid x_S}[\cdot]$. There are three common ways to obtain these distributions. (i) \emph{Direct marginals.} A learned model outputs $p(\mathbf{y}\mid x_S)$ and $p(\mathbf{x}_a\mid x_S)$ directly for arbitrary subsets $S$ and all $a \in U$. (ii) \emph{Bayes filtering.} A learned model outputs $p(\mathbf{x}_a\mid x_S,\mathbf{y})$ for arbitrary subsets $S$ and all $a \in U$, together with a prior $p(\mathbf{y}\mid x_{S_0})$, where $S_0$ denotes features that are always observed at zero cost. The posterior $p(\mathbf{y}\mid x_S)$ is maintained via the Bayesian update in \eqref{eq:bayes} as more features are observed, and $p(\mathbf{x}_a\mid x_S)$ is obtained by marginalizing over $\mathbf{y}$. (iii) \emph{Joint modeling.} A learned model outputs $p(\mathbf{x}_U,\mathbf{y}\mid x_S)$ for arbitrary subsets $S$. The required distributions $p(\mathbf{y}\mid x_S)$ and $p(\mathbf{x}_a\mid x_S)$ are then derived by marginalization (see \eqref{eq:obs1}). 

In many implementations, the marginalization over $\mathbf{x}_U$ needed to interpret prediction under partial observability is handled implicitly. In particular, the predictor is trained on masked (partially observed) inputs so that its output approximates $p(\mathbf{y}\mid x_S) = \mathbb{E}_{\mathbf{x}_U \mid x_S}[p(\mathbf{y}\mid x_S,\mathbf{x}_U)]$ without explicitly computing the expectation. This often makes it sufficient for planning to work with these marginals, rather than an explicit representation of the full joint belief.


\paragraph{AFA as an MDP and computational intractability.}
A distinctive property of the AFA-POMDP is that the observed information $(S,x_S)$ is a sufficient statistic for the full observation history. Concretely, $(S,x_S)$ \emph{is} the history, it records exactly which features have been acquired and their realized values. Therefore, each $(S,x_S)$ induces a unique belief $p(\mathbf{x}_U,\mathbf{y}\mid x_S)$, so the optimal value in \eqref{eq:pomdp_bellman_concise} can be written as a function of $(S,x_S)$ alone, without explicitly passing the full belief $b(x_S)$. Consequently, one can define a fully observable MDP with state space $\{(S,x_S): S\subseteq[d],\,x_S\in\mathcal{X}_S\}$ and Bellman recursion given by \eqref{eq:pomdp_bellman_concise}, with the belief terms understood as functions of $(S,x_S)$.

The main difficulty is computational: the number of possible outcomes $(S,x_S)$ grows combinatorially with $d$ (and becomes effectively uncountable for continuous features), so exact dynamic programming is feasible only for very small problems or under strong structural assumptions. In addition, practical AFA methods couple two challenges: (i) learning or approximating the conditional and marginal quantities induced by $p(\mathbf{x},\mathbf{y})$ that are needed for planning, and (ii) solving the resulting AFA-POMDP (or its MDP view) given those quantities (see Section \ref{sec:methods} for details). Importantly, even if the required probabilistic components were given, optimal planning remains computationally intractable in general. This is proven in \citep{10.5555/1641503.1641516}, which studies optimal value of information in graphical models, where  AFA-POMDP can be understood as a special case of their adaptive setting. They show that it is intractable except for chain-structured models (and a few closely related low-complexity cases), where an optimal adaptive acquisition policy can be computed efficiently via dynamic programming. These considerations motivate the approximate planning and learning approaches surveyed in the remainder of the paper.


Finally, a central question is when $(S,x_S)$ itself admits a lower-dimensional \emph{sufficient statistic} that preserves all decision-relevant information, so that we do not need to learn or represent a separate value $V$ for every realized outcome $(S,x_S)$. In particular, when features are conditionally independent given the label, such a compressed sufficient statistic exists \citep{9414669}. We discuss this case next.

\paragraph{Sufficient statistic under conditional independence.}
Assume features are conditionally independent given the label (i.e., the Naive Bayes assumption), so $p(\mathbf{x}_a\mid x_S,y)=p(\mathbf{x}_a\mid y)$ for all $a\notin S$. In the AFA-POMDP this eliminates the dependency of the observation model on the realized values $x_S$: the predictive distribution for a candidate feature becomes $p(\mathbf{x}_a=o\mid x_S)=\mathbb{E}_{\mathbf{y}\mid x_S}[p(\mathbf{x}_a=o\mid \mathbf{y})]$, and the Bayes update in \eqref{eq:bayes} uses only the class-conditional likelihood $p(\mathbf{x}_a\mid \mathbf{y})$. Consequently, the decision-relevant content of $x_S$ is fully captured by the posterior $p(\mathbf{y}\mid x_S)$, meaning the information state admits the reduced sufficient statistic $(S,p(\mathbf{y}\mid x_S))$: any two partial observations $x_S$ and $x'_S$ such that $p(\mathbf{y} \mid x_S) = p(\mathbf{y} \mid x'_S)$ are equivalent for prediction and planning. While the problem remains intractable in general under this assumption \citep{10.5555/1641503.1641516}, it can greatly improve tractability since the value function no longer needs to be represented separately for every realization $x_S\in\mathcal{X}_S$, and it clarifies why conditional independence is a common simplifying assumption in AFA.

More broadly, this perspective suggests studying other distributional assumptions under which compact sufficient statistics exist. Several works enrich the state with learned distributions such as $p(\mathbf{x}_U,\mathbf{y}\mid x_S)$ or summaries thereof \citep{pmlr-v139-li21p, GhoshLan2023DiFA}, but typically justify these choices heuristically. The POMDP formulation provides a principled lens for identifying when such augmentations are sufficient (or approximately sufficient) for near-optimal planning.

\paragraph{Motivation for POMDP view of AFA.} Apart from the discussion above, the POMDP perspective is also useful because it highlights the \emph{belief dynamics} that is intrinsic to the sequential acquisition for AFA. This makes the information flow explicit and clarifies which probabilistic quantities must be estimated. For example, estimating the required conditional quantities (e.g., $p(\mathbf{x}_a \mid x_S)$ and $p(\mathbf{y} \mid x_S)$) for arbitrary $S$ is typically intractable. The POMDP formulation highlights that only the conditional quantities used along the observed trajectory need to be estimated. This will significantly improve tractability, since subsets $S$ that occur with very low probability can be ignored. 

In addition, the belief formulation connects AFA to a well-studied class of POMDPs for which the value function over beliefs is piecewise linear. This is shown to hold for the AFA-POMDP in \citep{9414669} (assuming discretized features). This enables efficient dynamic programming and point-based approximations that operate on a sampled set of reachable beliefs rather than on the full belief space \citep{10.5555/1622519.1622525}. In principle, one can make use of any (approximate) POMDP solver proposed in the POMDP literature, see, e.g., \citep{10.5555/1036843.1036906,10.5555/1622519.1622525,10.5555/1622673.1622690}.

\paragraph{Properties of the AFA-POMDP.}
The AFA-POMDP has \emph{mixed observability}. The agent observes which features have been acquired and their realized values, denoted by $(S, x_S)$, whereas the label and the remaining (unacquired) features, $(\mathbf{y}, \mathbf{x}_U)$, remain latent. AFA therefore belongs to the class of Mixed-Observability MDPs (MOMDPs), where the belief state only needs to track the latent components. Leveraging this factorization can substantially reduce the effective dimensionality of the planning state \citep{10.1177/0278364910369861,5671411}.

The AFA-POMDP is also an \emph{information-gathering} POMDP in a passive environment. Actions do not alter the underlying instance $(x,y)$, they only reveal additional information about it (when the acquired features are informative). This structure has been exploited in theoretical analyses of POMDPs \citep{10.5555/3618408.3618609}. Consequently, maximizing return in the AFA-POMDP aligns with reducing uncertainty about the hidden state, for example by decreasing the entropy of $p(\mathbf{y}\mid x_S)$. To capture this, the reward is naturally \emph{belief-dependent}, that is, it depends on the posterior $p(\mathbf{y},\mathbf{x}_U \mid x_S)$ (see \eqref{eq:reward}). In contrast, for a general POMDP, maximizing reward need not correspond to reducing uncertainty in the belief over states. POMDPs with belief-dependent rewards have been studied in other contexts, particularly in active perception for robotics and related settings \citep{NIPS2010_68053af2,9899480,10.1145/3583068,Walraven2025_InfoGatheringPOMDPs}. Many of these works further shape the objective so that intermediate actions are rewarded for reducing uncertainty, an approach also used in AFA \citep{kachuee2018opportunistic, pmlr-v139-li21p}. While reward shaping can mitigate sparsity, it can also encourage more myopic policies.

Finally, the explicit \texttt{STOP} action yields action-based termination. The resulting undiscounted objective is naturally interpreted as a stochastic shortest path problem. It also facilitates the use of goal-oriented POMDP methods that rely on proper termination and bounded cost-to-go guarantees \citep{10.5555/1619797.1619844,ijcai2018p662,10.5555/3524938.3525698}.

\section{Myopic vs. Non-Myopic AFA}
\label{sec:myopic_nonmyopic}

A policy can be derived from the value function $V_k$ in \eqref{eq:pomdp_bellman_concise} by defining the corresponding action-value function. For $a = \texttt{STOP}$, $Q_k(b(x_S),\texttt{STOP})=V_0(b(x_S))$ for all $k$. Then, for any $a\in \mathcal{F}_{B}(S)$,
\begin{equation}
\begin{aligned}
\label{eq:qstar_acquire}
&Q_k(b(x_S),a) = -\alpha c_a +\mathbb{E}_{\mathbf{x}_a\mid x_S}[V_{k-1}(b(x_S,a,\mathbf{x}_a))].
\end{aligned}
\end{equation}
An optimal (deterministic) $k$-step truncated policy is then
\begin{equation}
\label{eq:pi_star_from_q}
\pi_k(S,x_S)\in\arg\max_{a \in \mathcal{A}_B(S)} Q_k(b(x_S),a).
\end{equation}
All AFA methods effectively amount to approximating the action-value function $Q \approx Q_d$ (or directly the policy $\pi_d$ for policy-gradient methods) by making different modeling and computational choices. A natural approximation is $Q_k \approx Q_d$ for $k < d$. Finally, for $a \in \mathcal{F}_B(S)$, we denote 

\begin{equation}
\Delta_k(x_S, a) \triangleq Q_k(b(x_S),a) - Q_k(b(x_S),\texttt{STOP}),
\end{equation}

which represents the benefit of acquiring feature $a$ over stopping in the current belief state $b(x_S)$ under a $k$-step truncation. Let $a^\ast = \arg\max_{a \in \mathcal{F}_B(S)}\Delta_k(x_S, a)$. Then, if $\Delta_k(
x_S, a^\ast) \leq 0$ we stop the acquisition process, and acquire feature $a^\ast$ otherwise.

\paragraph{Myopic heuristics.} 

The most myopic approximation corresponds to the $1$-step truncated action-value function $Q_1$. For any $a\in \mathcal{F}_B(S)$,
\begin{equation}
\begin{aligned}
\label{eq:q_myop_acquire}
&Q_1(b(x_S),a)= -\alpha c_a + \mathbb{E}_{\mathbf{x}_a\mid x_S}[V_0(b(x_S,a,\mathbf{x}_a))],
\end{aligned}
\end{equation}
where $V_0(b(x_S,a,\mathbf{x}_a) = r(b(x_S,a,\mathbf{x}_a),\texttt{STOP})$ (defined in Eq. \ref{eq:reward}). Computing $Q_1$ requires no planning beyond the marginalization over $\mathbf{x}_a$. If we assume the reward in \eqref{eq:reward} is defined in terms of log loss, we have

\begin{equation} \label{eq:CMI}
\begin{aligned}
\Delta_1(x_S, a) &=  H(\mathbf{y} \mid x_S) - \mathbb{E}_{\mathbf{x}_a\mid x_S}[H(\mathbf{y}\mid \mathbf{x}_a,x_S)] -\alpha c_a \\
&= I(\mathbf{y};\mathbf{x}_a \mid x_S) -\alpha c_a. 
\end{aligned}
\end{equation}

$I(\mathbf{y};\mathbf{x}_a \mid x_S)$ is the conditional mutual information (CMI) between $\mathbf{x}_a$ and $\mathbf{y}$. The myopic CMI policy is one of the most common heuristics in the AFA literature (often in combination with the Naive Bayes assumption), and often performs well \citep{schütz2025afabenchgenericframeworkbenchmarking}. Other variants of this heuristic exist. For example, the literature on adaptive submodularity suggests $I(\mathbf{y};\mathbf{x}_a\mid x_S)/c_a$) \citep{golovin2011adaptive}. More broadly, the myopic approximation is connected to the myopic value of information from decision theory and other related fields \citep{f1168103-f0ea-30c5-aa6c-babe695713d3, 4082064, 10.5555/1641503.1641516}. 


\begin{table*}[t]
\centering
\setlength{\tabcolsep}{5pt}
\renewcommand{\arraystretch}{1.15}
\small
\begin{tabular}{p{0.28\textwidth} p{0.68\textwidth}}
\toprule
\textbf{Category} & \textbf{Papers} \\
\midrule

\textbf{Embedded Cost-Aware Predictors} & \\
\quad Cost-sensitive decision trees
& \citep{Norton1989,Tan1990CSL,Nunez1991EG2,10.5555/1622826.1622838,ling2004decision,10.5555/1597538.1597616,1613866, DBLP:conf/icml/ShengL06, 1644729} \\
\quad Cost-sensitive ensembles
& \citep{miser2012,2015forest,nan2016pruning, NIPS2017_4fac9ba1} \\
\midrule
\textbf{Model-Based Methods} & \\
\quad Myopic generative heuristics
& \citep{geman1996active,chai2004test, DBLP:journals/pr/JiC07, nan2014fast, ma2019eddi, NEURIPS2019_c055dcc7, RangrejC21, 10.1145/3430984.3431008, chattopadhyay2022interpretable, bsoda2022, mirzaei2023fast} \\
\quad Non-myopic generative heuristics
& \citep{DBLP:conf/aaai/BilgicG07,DBLP:journals/jair/ChenCD14,uaivoi2017, karanam2023on, beebe-wang2023explanationguided, DBLP:conf/icml/ValanciusLO24,ijcai2023p463,rahbar2025costefficient,huang2025informationtemplatesnewparadigm, norcliffe2025stochasticencodingsactivefeature} \\
\quad Dynamic programming
& \citep{Maliah_Shani_2018,9422113,9414669} \\
\quad Search
& \citep{10.5555/645531.656020, BLP:conf/uai/Zubek04, DBLP:conf/icdm/ArntZ04, DBLP:journals/jair/ZubekD05} \\

\midrule
\textbf{Model-Free Methods} & \\
\quad Myopic discriminative heuristics
& \citep{DBLP:conf/nips/HeDE12,covert2023learning,GhoshLan2023DiFA,chattopadhyay2023variational,gadgil2024estimating} \\
\quad Non-myopic discriminative heuristics
& \citep{DBLP:journals/corr/HeMK16} \\
\quad Model-free RL
& \citep{dulac2011datum, DBLP:journals/ml/Dulac-ArnoldDPG12, 10.1007/978-3-642-25832-9_14, ruckstiest2013minimizing,NEURIPS2018_e5841df2,kachuee2018opportunistic, 10.1609/aaai.v33i01.33013959, DBLP:journals/ml/JanischPL20} \\
\midrule
\textbf{Hybrid Methods} & \\
\quad Myopic hybrid heuristics
& See discussion in Section \ref{sec:hybrid}. \\
\quad Non-myopic hybrid heuristics
& \citep{GhoshLan2023DiFA, DBLP:conf/icml/ValanciusLO24, huang2025informationtemplatesnewparadigm, guney2025active}. \\
\quad Model-based RL
& \citep{zannone2019odin, 9533593, pmlr-v139-li21p, li2024distributionguidedactivefeature} \\
\bottomrule
\end{tabular}
\caption{Taxonomy of AFA methods. See Section \ref{sec:methods} for details.}
\label{tab:afa_methods_taxonomy}
\end{table*}

\paragraph{Non-myopic AFA.}

The optimal action-value function $Q_d$ and the myopic heuristic $Q_1$ are at the two extreme ends of policies for AFA. However, it is often of interest to find a policy between these extreme ends that finds a balance between (i) computation time, and (ii) performance. To understand what makes a policy non-myopic for AFA (i.e., planning beyond a one-step lookahead), we consider the two-step truncated action-value function below (assuming log loss). For $a \in \mathcal{F}_{B}(S)$, we have

%

\begin{equation}
\label{eq:delta2}
\begin{aligned}
&Q_2\big(b(x_S),a\big)
= -\alpha c_a +
\mathbb E_{\mathbf x_a\mid x_S}\Big[
\max \Big\{-H(\mathbf y\mid x_S,\mathbf x_a), \\
&\max_{b\in \mathcal{F}_B(S \cup \{a\})}
(
-\alpha c_b
+
\mathbb E_{\mathbf x_b\mid x_S,\mathbf x_a}[-H(\mathbf y\mid x_S, \mathbf x_a,\mathbf x_b)]
)
\Big\}
\Big].
\end{aligned}
\end{equation}

Here, $I(\mathbf{y};\mathbf{x}_b\mid x_S,\mathbf{x}_a)$ is the conditional mutual information between $\mathbf{y}$ and $\mathbf{x}_b$ after observing $\mathbf{x}_a$. This makes it explicit that non-myopic planning must account for interactions between a candidate feature $a\in U$ and remaining unobserved features in $U\setminus\{a\}$. For example, two features $a\neq b$ can each be individually uninformative (large $H(\mathbf{y} \mid x_S, \mathbf{x}_a)$ and $H(\mathbf{y} \mid x_S, \mathbf{x}_b)$) yet be jointly decisive, meaning $H(\mathbf{y} \mid x_S,\mathbf{x}_b,\mathbf{x}_a)$ can be very small. Clearly, this effect can not be captured by the myopic heuristic in \eqref{eq:q_myop_acquire}. 

\paragraph{Non-myopic acquisitions via a non-adaptive approximation.} 

In \eqref{eq:delta2}, we see that in the expectation $\mathbb{E}_{\mathbf{x}_a \mid x_S}[\cdot]$, we consider the value of the optimal next action for each outcome of $\mathbf{x}_a$. It is primarily this \emph{adaptive} nature of AFA that makes it highly intractable. Hence, an important approximation is to move the maximization over actions outside the expectation. The next proposition presents this, where $\mathcal{N}_B(S,k, a) \triangleq \{O \subseteq [d] \setminus S : a \in O, c(S \cup O) \leq B, |O| \leq k\}$.

\begin{proposition} \label{prop:nonadaptive}
If we move the maximization over actions outside the expectation $\mathbb{E}_{\mathbf{x}_a \mid x_S}[\cdot]$ in \eqref{eq:delta2}, the value of action $a \in \mathcal{F}_B(S)$ is $Q_2^{\text{NA}}(b(x_S),a) \triangleq \max_{O \in \mathcal{N}_B(S,2,a)}-H(\mathbf{y} \mid x_S,\mathbf{x}_O) - c(O)$.
\end{proposition}

Moreover, we can generalize Proposition \ref{prop:nonadaptive} to arbitrary truncation $k$. Then, the benefit of selecting $a \in \mathcal{F}_B(S)$ over stopping is:

\begin{equation} \label{eq:nonadaptive}
\begin{aligned}
    \Delta_k^{\text{NA}}(x_S,a) &\triangleq Q_k^{\text{NA}}(b(x_S),a) - Q_k^{\text{NA}}(b(x_S),\texttt{STOP}) \\
    &= \max_{O \in \mathcal{N}_B(S,k,a)}I(\mathbf{y};\mathbf{x}_O \mid x_S) - \alpha c(O).
\end{aligned}
\end{equation}

We can remove the truncation by setting $k = d$. Note that $\Delta_k^{\mathrm{NA}} = \Delta_k$ when $k=1$, but this equality does not hold in general for $k \ge 2$. The optimization problem in \eqref{eq:nonadaptive} is a constrained set optimization problem. With non-uniform costs, the constraint in \eqref{eq:nonadaptive} is a knapsack constraint, whereas with uniform costs it reduces to a cardinality constraint \citep{DBLP:books/cu/p/0001G14}. Importantly, \eqref{eq:nonadaptive} amounts to (cost-aware) static feature selection \citep{DBLP:journals/jmlr/GuyonE03, JMLR:v13:brown12a} conditioned on $x_S$. Thus, at the current belief state $b(x_S)$, we solve a non-adaptive feature selection problem as an approximation to the adaptive nature of AFA. The problem in \eqref{eq:nonadaptive} is intractable in general \citep{DBLP:journals/jmlr/GuyonE03}. However, because it is non-adaptive, one can leverage established results from submodular set function optimization to obtain high-quality solutions. In particular, when the objective is \emph{monotone} and \emph{submodular}, a greedy algorithm yields a provably near-optimal solution to \eqref{eq:nonadaptive} \citep{10.5555/3020336.3020377, DBLP:books/cu/p/0001G14}. Moreover, $I(\mathbf{y}; \mathbf{x}_O \mid x_S)$ is monotone and submodular under conditional independence of features given the label \citep{10.5555/3020336.3020377}. In addition, a number of other heuristic approaches have been proposed to solve \eqref{eq:nonadaptive} in the literature on static feature selection, see e.g., \citep{JMLR:v13:brown12a} (for example, by only considering pairwise feature interactions and ignoring higher order interactions). 

Let $O^\ast$ be the set that maximizes \eqref{eq:nonadaptive} across all $a \in \mathcal{F}_B(S)$. Then, all $a \in O^\ast \subseteq \mathcal{F}_B(S)$ are assigned an equal value. Since we ultimately acquire a single feature, we therefore require a tie-breaking rule. One simple option is to select, among the tied actions, a feature $a \in O^\ast$ with the largest marginal mutual information $I(\mathbf{y}; \mathbf{x}_a \mid x_S)$. Alternatively, one can select the feature with the largest expected conditional contribution to the joint informativeness, namely $\mathbb{E}_{\mathbf{x}_{O^\ast \setminus \{a\}}} [I(\mathbf{y}; \mathbf{x}_a \mid x_S, \mathbf{x}_{O^\ast \setminus \{a\}})] = I(\mathbf{y}; \mathbf{x}_{O^\ast} \mid x_S) - I(\mathbf{y}; \mathbf{x}_{O^\ast \setminus \{a\}} \mid x_S)$. 

In fact, a possible (but substantially more naive) selection rule is to choose the feature that maximizes $\mathbb{E}_{\mathbf{x}_{U \setminus \{a\}}}[I(\mathbf{y}; \mathbf{x}_a \mid \mathbf{x}_S, \mathbf{x}_{U \setminus \{a\}})]$, where $U$ denotes the set of all remaining unobserved features. The underlying idea is to score each candidate feature by averaging its utility over \emph{imputations} of the missing features. While this type of heuristic has been considered in previous work on AFA \citep{beebe-wang2023explanationguided, karanam2023on, norcliffe2025stochasticencodingsactivefeature}, it is still a naive way to approximate non-myopic selection, as it ignores feature acquisition costs.

Comparing the myopic CMI objective $I(\mathbf{y}; \mathbf{x}_a \mid x_S)$ with its non-myopic counterpart
$\mathbb{E}_{\mathbf{x}_{U \setminus \{a\}}} \left[I(\mathbf{y}; \mathbf{x}_a \mid \mathbf{x}_S, \mathbf{x}_{U \setminus \{a\}})\right]$
also further clarifies the conceptual distinction between myopic and non-myopic selection for AFA. In the myopic objective, one first marginalizes over the unobserved features and then evaluates the acquisition score. Concretely, one forms
$p(\mathbf{y} \mid x_S) = \mathbb{E}_{\mathbf{x}_U} \left[p(\mathbf{y} \mid x_S, \mathbf{x}_U)\right]$,
and this marginalized predictive distribution is then used \emph{inside} $I(\mathbf{y}; \mathbf{x}_a \mid x_S)$.
In contrast, the non-myopic objective evaluates the information gain w.r.t. $p(\mathbf{y} \mid x_S, x_U)$ in expectation over outcomes of $x_U \sim p(\mathbf{x}_U \mid x_S)$ (i.e., the marginalization happens \emph{outside} of the mutual information).

In batch selection variants of AFA \citep{DBLP:conf/icml/ShengL06}, where multiple features are acquired at each selection step, one could solve \eqref{eq:nonadaptive} and acquire all features in the optimal set $O^\ast$, where $k$ is then interpreted as the batch size. Batch selection is well studied in previous work on adaptive information acquisition (beyond AFA) \citep{pmlr-v28-chen13b, 10.1145/3472291}.

The idea of performing static feature selection at the current belief state $b(x_S)$ has appeared in prior work on AFA beyond mutual information objectives. For example, existing approaches have used LIME/SHAP-based attributions \citep{beebe-wang2023explanationguided, guney2025active} as well as gradient-based feature importance scores \citep{norcliffe2025stochasticencodingsactivefeature}. Finally, \citep{DBLP:conf/icml/ValanciusLO24, huang2025informationtemplatesnewparadigm} study $Q^{\mathrm{NA}}_k$ for AFA (although they do not connect it to static feature selection as we do here) and show that $Q^{\mathrm{NA}}_k \leq Q_k$. A key direction for future work is to characterize the quality of the approximation $Q^{\mathrm{NA}}_k \approx Q_k$. For instance, it would be valuable to determine whether $Q^{\mathrm{NA}}_k$ can be modified to yield a tighter lower bound on $Q_k$. The following discussion may provide some ideas in this direction.

\paragraph{Near-optimality guarantees via adaptive submodularity.}
Above, we discussed how near-optimality guarantees for the non-adaptive problem in \eqref{eq:nonadaptive} can be obtained via submodularity. Analogous guarantees extend to adaptive decision making through \emph{adaptive submodularity} \citep{golovin2011adaptive}. While \citep{golovin2011adaptive} studies general \emph{adaptive stochastic optimization} problems, AFA can be viewed as a special case. 

To build intuition, let $u(b(x_S))$ denote the utility of the current belief state $b(x_S)$, and define the average utility of a policy $\pi$ under budget $B$ as $u_{\mathrm{avg}}(\pi,B) = \mathbb{E}_{\mathbf{x},\mathbf{y}}\!\left[u\!\left(b(\mathbf{x}_{\pi[\mathbf{x}]})\right)\right]$, with $\pi^\ast$ denoting the policy that maximizes $u_{\mathrm{avg}}(\pi,B)$. Let $\pi_{\mathrm{greedy}}$ be the policy that, at each step, selects a feature $a \in \mathcal{F}_B(S)$ maximizing the expected one-step improvement relative to the cost, i.e., $(u(b(x_S)) - \mathbb{E}_{\mathbf{x}_a \mid x_S}[u(b(x_S,a,\mathbf{x}_a))])/c_a$. If $u$ satisfies \emph{adaptive monotonicity} and \emph{adaptive submodularity} (see \citep{golovin2011adaptive} for formal definitions), then $u_{\mathrm{avg}}(\pi_{\mathrm{greedy}},B) \geq (1-\frac{1}{e})\,u_{\mathrm{avg}}(\pi^\ast,B)$, meaning that the greedy policy achieves a constant-factor approximation to the optimal policy. For the choice $u(b(x_S)) = \ell(f(x_S),y) + \alpha c(S)$, $u_{\mathrm{avg}}(\pi,B)$ coincides with the AFA objective in \eqref{eq:afa_obj_soft}. In particular, under log loss and $\alpha=0$, we have $u(b(x_S)) = -H(\mathbf{y}\mid x_S)$, and $\pi_{\mathrm{greedy}}$ reduces to the myopic CMI policy in \eqref{eq:CMI}. However, $-H(\mathbf{y}\mid x_S)$ is known not to be adaptive submodular, even under a naive Bayes assumption \citep{chen2015sequential}, and adaptive submodularity also fails under $0$--$1$ loss \citep{golovin2010near}. This aligns with the inherently myopic nature of \eqref{eq:CMI}. Still, \citep{chen2015sequential} shows that the myopic CMI policy in \eqref{eq:CMI} can achieve near-optimal performance under restrictive assumptions on the data distribution $p(\mathbf{x},\mathbf{y})$.

The only objective suited for AFA that is known to be adaptive submodular is $\mathrm{EC}^2$, introduced in \citep{golovin2010near} (and direct extensions of $\mathrm{EC}^2$, see \citep{chen2017phd}). In the AFA literature, $\mathrm{EC}^2$ has been used in \citep{ijcai2023p463, rahbar2025costefficient}. The $\mathrm{EC}^2$ objective is designed so that greedy optimization induces non-myopic behavior for AFA, in the sense that it approximates the multi-step lookahead value $Q_k$ for $k>1$. However, $\mathrm{EC}^2$ is connected to, but not fully aligned with, the AFA objective in \eqref{eq:afa_obj_soft}. Thus, $\mathrm{EC}^2$ can be viewed as a surrogate objective: greedy adaptive optimization enjoys near-optimality guarantees for $\mathrm{EC}^2$, but the resulting policy may be suboptimal for \eqref{eq:afa_obj_soft} when the two objectives diverge. We briefly introduce $\mathrm{EC}^2$ in Section \ref{sec:methods}.


\paragraph{Online planning in the AFA-POMDP.} 
For general POMDPs, when exact planning is intractable, it is common to rely on online planning from the current belief state \citep{10.5555/1622673.1622690}. A widely used approach is \emph{Monte-Carlo tree search} (MCTS) \citep{10.5555/1777826.1777833, NIPS2010_edfbe1af, browne2012mcts}, which was applied to AFA in \citep{9533593}. For AFA, MCTS performs repeated rollouts of candidate future action sequences from the current belief using a rollout policy, aggregates the resulting returns to expand a lookahead tree, and selects the next action that appears best on average. 

Moreover, many AFA methods that exploit $p(\mathbf{x}_U \mid x_S)$ to make non-myopic feature acquisition decisions at each belief state $b(x_S)$ can be interpreted as online planning in the underlying AFA-POMDP. This perspective includes both the non-adaptive approximation $Q^{\mathrm{NA}}_k$ and the $\mathrm{EC}^2$ objective discussed above.

\section{AFA Methods} \label{sec:methods}

As discussed in  Section~\ref{sec:problem_formulation}, AFA is naturally a POMDP, and existing methods mainly differ in how they approximate the optimal action-value in \eqref{eq:qstar_acquire}. We group them into three categories: (1) \emph{embedded cost-aware predictors}, where the model structure implicitly defines an acquisition trajectory; (2) \emph{model-based methods}, which estimate the probabilistic quantities of the POMDP (e.g., $p(\mathbf{y}\mid x_S)$ and $p(\mathbf{x}_a\mid x_S)$) and then find a policy via model-based planning; (3) \emph{model-free}, which learn a policy directly from acquisition rollouts, and (4) \emph{hybrid} methods which learn a policy from acquisition rollouts in combination with a learned model to improve stability. See Table \ref{tab:afa_methods_taxonomy} for a taxonomy of all papers. This taxonomy also parallels standard POMDP solution paradigms \citep{10.5555/1622673.1622690, 9899480} (often not stated in AFA papers). Some approaches are primarily \emph{offline} (they learn or precompute a policy from training data or a learned model), while others are \emph{online} (they plan at test time from the current belief $b(x_S)$). In addition, some are myopic while others are non-myopic by planning over a longer future horizon. 

Finally, most methods focus on one of the two special cases: $\alpha = 0$ or $B = c([d])$. This corresponds to a hard budget and a soft budget, respectively. See the discussion in Section~\ref{sec:problem_formulation}. In addition, some methods train predictor $f$ and policy $\pi$ separately, while some train them jointly (as discussed in Section \ref{sec:problem_formulation}). However, for brevity, we do not always state these distinctions for each method.

\subsection{Embedded Cost-Aware Predictors}
\label{sec:embedded}

Any deterministic AFA policy $\pi$ can be represented as a decision tree \citep{quinlan1993c4.5}. At test time, a decision tree evaluates only a subset of features, thereby implementing adaptive, instance-wise feature acquisition within the model itself. Consequently, decision-tree learners that explicitly incorporate feature costs yield valid policies for AFA. This section covers such cost-aware decision tree methods. All methods in this section also fall under one of the broader (POMDP-based) categories introduced later, since they can be interpreted as approximations to the underlying POMDP for AFA (as any AFA policy can). We nevertheless present them as a separate category to emphasize that they are specific to decision tree learning. Moreover, in contrast to other AFA methods introduced later, they inherit familiar limitations of standard decision trees (e.g., overfitting and discretizing continuous features). They are rarely included in modern AFA benchmarks.

\paragraph{Cost-sensitive decision trees and ensembles.}
A standard decision tree is typically constructed myopically. For example, it decides which feature to split based on the feature with largest information gain $I(\mathbf{y};\mathbf{x}_a\mid x_S)$ \citep{quinlan1993c4.5}. An early example of a cost-sensitive decision tree is \citep{Norton1989}, where the split criteria now takes the cost of the feature into account as $I(\mathbf{y};\mathbf{x}_a\mid x_S)/c_a$ (i.e., connected to the myopic CMI heuristic in \eqref{eq:CMI}). This requires access to $p(\mathbf{x}_a \mid x_S)$ and $p(\mathbf{y} \mid x_S)$, which are estimated from the empirical frequencies of the training examples that reach the current node (i.e., consistent with $x_S$). This is a form of generative estimation of these probabilities. These methods hence also belong to the category \emph{myopic generative heuristics} in Section \ref{sec:model_based_methods}. Other methods are similar in spirit, but consider other myopic cost-sensitive split criteria \citep{Tan1990CSL, Nunez1991EG2, 10.5555/1622826.1622838, DBLP:conf/icml/ShengL06,10.5555/1597538.1597616,1644729,1613866}. Some of these methods also consider non-myopic variants via lookahead during splits \citep{Norton1989} or post-processing of the myopic tree by optimizing some global objective (e.g., \eqref{eq:afa_obj_soft}) \citep{10.5555/1622826.1622838}. This aligns them with one of the \emph{non-myopic heuristic} categories discussed later. Notably, \citep{1644729} rebuilds a cost-sensitive decision tree at test time after each feature is acquired, with cost of already acquired features set to zero. This corresponds to a kind of online planning in the underlying AFA-POMDP. 

This idea has also been extended to ensemble methods, e.g, random forests \citep{miser2012, 2015forest, nan2016pruning} and boosting methods \citep{NIPS2017_4fac9ba1}. The general idea is to build multiple cost-sensitive decision trees, and combine them similar to standard ensemble methods. A feature is only acquired the first time it is used by a tree in the ensemble.



\subsection{Model-Based Methods} \label{sec:model_based_methods}

Model-based methods explicitly estimate $p(\mathbf{x}_U \mid x_S)$ (or marginals of this) for arbitary $S$ in order to perform model-based planning in the AFA-POMDP. While these methods also require access to $p(\mathbf{y} \mid x_S)$, this dependence is not specific to model-based approaches. As discussed in Section \ref{sec:pomdp_formulation}, model-free methods (covered in Section \ref{sec:model_free}) also require access to $p(\mathbf{y} \mid x_S)$ (or, more generally, a predictor $f(x_S)$) to evaluate the stopping reward $R(s(x_S,y),\texttt{STOP})$ in \eqref{eq:reward}. The main differences between model-based methods are (i) how they approximate $p(\mathbf{x}_U \mid x_S)$ and $p(\mathbf{y} \mid x_S)$, and (ii) how they use these estimates to approximate the action-value function in \eqref{eq:qstar_acquire}.


In addition, there are two primary approaches for computing the probabilities required for model-based planning at a belief state $b(x_S)$: (i) \emph{Parametric (deep) generative models.} Here, one fits a (potentially deep) parametric model and uses it to estimate the required conditional probabilities given $x_S$. (ii) \emph{Non-parametric, neighborhood-based estimation.} Here, one explicitly identifies a neighborhood of training samples around $x_S$. For example, this can be done using a $k$-nearest neighbor ($k$-NN) approach to find training samples that are close to $x_S$ on features in $S$. Then, the required conditional quantity (given $x_S$) is estimaed using the empirical frequencies within this neighborhood. The motivation for the neighborhood approach is typically to avoid the training instability in the methods based on deep generative models (especially for complex datasets) \citep{DBLP:conf/icml/ValanciusLO24}.

\paragraph{Myopic generative heuristics.}
The methods in this category compute the myopic approximation in \eqref{eq:q_myop_acquire}, with different choices of loss $\ell$ in the terminal reward $r(b(x_S), \texttt{STOP})$ in \eqref{eq:reward}. The earliest methods in this category are the myopic cost-sensitive decision trees discussed in Section \ref{sec:embedded}. We now discuss other approaches, not based on decision trees. Early methods include \citep{geman1996active, chai2004test}, both of which assume a naive Bayes setting. They maintain the required probabilities at inference via Bayesian updating (see \eqref{eq:bayes}) and consider log loss and 0-1 loss in the reward \eqref{eq:reward}, respectively. \citep{DBLP:journals/pr/JiC07} also use the 0-1 loss, but goes beyond the naive Bayes assumption by considering richer class-conditional generative models (e.g., IOHMM/HMM beliefs to compute both $p(\mathbf{y}\mid x_S)$ and $p(\mathbf{x}_a\mid x_S)$). They also use Bayesian updating to maintain probabilities.

In \citep{nan2014fast}, a fast margin-based method is proposed. Given observed features $S$, it forms a $k$-NN neighborhood of the test instance in the observed subspace and estimates confidence as the fraction of neighboring training points with positive partial margin (equivalently, correctly classified by the partial classifier). It then selects the next feature to maximize the resulting increase in this neighborhood-based positive-margin fraction (minus a feature-cost penalty).

More recently, deep models have been used to estimate these probabilities without making any simplifying assumptions about $p(\mathbf{x}, \mathbf{y})$. This is implemented using \emph{deep arbitrary conditional generative} models to approximate the required conditional probabilities. A major limitation of these methods is that they aim to estimate probabilities for arbitrary feature subsets $S$, rather than restricting attention to the subsets that are likely to arise at inference time. This mismatch can make training substantially more difficult and can also degrade AFA performance \citep{schütz2025afabenchgenericframeworkbenchmarking}.

\citep{ma2019eddi} propose a novel \emph{partial variational autoencoder} (Partial VAE) to achieve this. Then, they compute $I(\mathbf{y};\mathbf{x}_a \mid x_S)$ (myopic CMI rule, see \eqref{eq:CMI}) given this. In \cite{chattopadhyay2022interpretable}, a similar approach has been used to implement the CMI policy, which also uses a VAE together with an MCMC algorithm. \citep{RangrejC21} adapts the partial VAE proposed in \citep{ma2019eddi} to perform predictions on images. \cite{NEURIPS2019_c055dcc7} proposes PA-BELGAM, a Bayesian extension of the partial VAE model in \citep{ma2019eddi} (which also allows them to perform active acquisition of features at training).

A limitation of EDDI is the computational cost involved in computing $I(\mathbf{y};\mathbf{x}_a \mid x_S)$ given the partial VAE at inference (it often requires Monte-Carlo sampling). In \cite{bsoda2022}, a Product-of-Experts (PoE) encoder \citep{wu2018multimodal} is employed instead of the partial VAE. The PoE encoder leverages the final predictor model to approximate the posterior distribution of the latent variable. Utilizing the PoE encoder for posterior inference allows for a reduction in the computational cost of estimating CMI.

There are a number of additional studies on deep arbitrary conditional models \citep{ivanov2018variational, molina2019spfloweasyextensiblelibrary, NEURIPS2019_5a0c8283, pmlr-v119-li20a, strauss2021arbitrary}, but not necessarily in the context of AFA. 

\citep{10.1145/3430984.3431008} assume that a subset of cost-free features, denoted $S_0$, is always available. They first cluster the data points using only the features in $S_0$. Next, for each cluster, they identify the most informative additional features via a static feature selection method. At test time, given a new instance $x$, they assign it to a cluster based on $x_{S_0}$ and then acquire the features deemed most informative for that cluster.

\citep{mirzaei2023fast} identify a neighborhood of $x_S$ within the training set, compute feature informativeness over this neighborhood using ANOVA $F$-scores (a simple univariate, static feature selection criterion), and then acquire the single most informative feature.

\paragraph{Non-myopic generative heuristics.} The methods in this category are non-myopic because they consider the unobserved features beyond a one-step lookahead. However, they do not use standard POMDP planning to achieve this. Instead, they are non-myopic by exploiting the specific structure of the AFA problem (i.e., by estimating the joint informativeness of unobserved features, see discussion in Section \ref{sec:myopic_nonmyopic}). 

\citep{DBLP:conf/aaai/BilgicG07} estimates the joint informativeness of feature groups and then selects a single feature from the most informative group (i.e., similar to $Q^{\text{NA}}_k$ in \eqref{eq:nonadaptive}). To make this tractable, it assumes conditional independence among features.

\citep{golovin2011adaptive} introduced the \(\mathrm{EC}^2\) objective in a more general active learning setup, which was later used for AFA in \citep{uaivoi2017, ijcai2023p463, rahbar2025costefficient}. \(\mathrm{EC}^2\) assigns each instance $x \in \mathcal{X}$ to a decision region $\mathcal{R}_y$ for each $y \in \mathcal{Y}$, where $\mathcal{R}_y$ contains the instances for which label $y$ is most likely (with respect to a chosen loss). Given a belief state $b(x_S)$, it constructs a graph whose nodes correspond to instances, and whose edges connect pairs of instances $(x,x')$ that lie in \emph{different} decision regions and remain consistent with the current observation $x_S$. Each such edge is weighted by $p(x)p(x')$. The next feature is chosen to maximize the expected reduction in total edge weight, normalized by the feature cost. This objective is \emph{adaptive submodular}, so greedy selection is near-optimal w.r.t. minimizing the \(\mathrm{EC}^2\) objective. However, this objective may not be aligned with the AFA objective in \eqref{eq:afa_obj_soft}, see Section~\ref{sec:myopic_nonmyopic} for further discussion.

\citep{DBLP:journals/jair/ChenCD14} introduce the same-decision probability (SDP). The SDP of \(x_S\) is the probability (in expectation over the unobserved features $\mathbf{x}_U$) that the label predicted from \(x_S\) matches the label predicted when all features in \(U\) are revealed. SDP selects the feature that maximizes the increase in this probability. It is intractable in general, but can be efficiently computed in certain graphical models \citep{DBLP:journals/jair/ChenCD14}.

\citep{karanam2023on} proposes to acquire the feature that maximally reduces $\mathbb{E}_{\mathbf{x}_U \mid x_S}[f(x_S,\mathbf{x}_U)^2]-
(\mathbb{E}_{\mathbf{x}_U \mid x_S}\left[f(x_S,\mathbf{x}_U)\right])^2$ in expectation. This corresponds to average the variance of the output of the predictor at the current belief state $b(x_S)$ across unobserved features in $U$. They estimate $p(\mathbf{x},\mathbf{y})$ using a probabilistic circuit (PC) \citep{ProbCirc20}. A PC represents a joint distribution as a sum and product computation graph. Under standard structural constraints (for example smoothness and decomposability), PCs support efficient marginalization and conditioning (such that required distributions conditioned on $x_S$ can be computed).

\citep{beebe-wang2023explanationguided} applies SHAP-based feature importances to AFA. At a belief state $b(x_S)$, they identify a set of training instances similar to $x_S$ using $k$-NN. They then compute \emph{local} SHAP attribution scores for all features on each instance in this set, and acquire the currently unobserved feature with the largest mean attribution across the instances in this set. In other words, they approximate the epxectation over $\mathbf{x}_U$ using $k$-NN. The method of \citep{norcliffe2025stochasticencodingsactivefeature} is similar in spirit, but replaces SHAP with gradient-based local attributions, and uses a deep generative model to estimate the expectation over $\mathbf{x}_U$ in a latent space. These methods are similar in spirit to $\mathbb{E}_{\mathbf{x}_{U \setminus \{a\}}} [I(\mathbf{y}; \mathbf{x}_a \mid x_S, \mathbf{x}_{U \setminus \{a\}})]$ discussed in Section \ref{sec:myopic_nonmyopic} (where SHAP/gradient-based feature attributions are used in place of the mutual information).

Finally, the methods of \citep{DBLP:conf/icml/ValanciusLO24, huang2025informationtemplatesnewparadigm} explicitly compute $Q^{\text{NA}}_k$ in \eqref{eq:nonadaptive} for $k = d$. That is, they rely on static feature-importance scores while explicitly incorporating feature costs, see the discussion surrounding \eqref{eq:nonadaptive} in Section~\ref{sec:myopic_nonmyopic}. The two approaches are closely related and mainly differ in how they optimize \eqref{eq:nonadaptive}. \citep{DBLP:conf/icml/ValanciusLO24} uses a simple Monte Carlo strategy based on randomly sampled feature subsets, whereas \citep{huang2025informationtemplatesnewparadigm} extends this framework by identifying better subsets via an offline set-optimization problem (leveraging the submodularity of entropy under naive Bayes, see Section~\ref{sec:myopic_nonmyopic}). Finally, they also use a $k$-NN approach to approximate the expectation over the subset of unobserved features $\mathbf{x}_O$.

\paragraph{Dynamic programming and search.}
\citep{9414669} proposes to solve the POMDP of AFA via dynamic programming. To make it tractable, they utilize (i) the piecewise-linear structure of the value function of the AFA-POMDP, and (ii) the reduced sufficient statistic via conditional independence as described in Section \ref{sec:pomdp_formulation}. \citep{9422113} uses dynamic programming in a related manner, but studies a simplified AFA setting in which the feature acquisition order is fixed. As a result, the policy is adaptive only in deciding whether to continue acquiring additional features or to stop and make a prediction. This restriction substantially improves tractability, but it is also highly limiting in practice. An alternative is to discretize the information state via learned abstractions, for example using the leaves of decision trees as surrogate states \citep{Maliah_Shani_2018}.

Moreover, the state space of the AFA-POMDP corresponds to an AND-OR graph. \citep{DBLP:journals/jair/ZubekD05} proposes to search for a policy in this graph using the AO$^\ast$ serach algorithm. They propose an admissible heuristic and statistical pruning to make it tractable. \citep{DBLP:journals/jair/ZubekD05} is a direct extension of \citep{10.5555/645531.656020, BLP:conf/uai/Zubek04}.

\subsection{Model-Free Methods}
\label{sec:model_free}

Model-free methods sidestep explicit probabilistic modeling of transition and observation dynamics (i.e., $p(\mathbf{x}_U \mid x_S)$), and instead learn acquisition behavior directly from simulating episodes on fully observed data instances $(x, y) \sim p(\mathbf{x}, \mathbf{y})$. Rather than estimating probabilities and plugging them into an acquisition objective, they learn to predict the value of actions in a given state (e.g., via an action-value function), typically using a neural network. As discussed (see Sections \ref{sec:pomdp_formulation} and \ref{sec:model_based_methods}), model-free methods still require access to $p(\mathbf{y} \mid x_S)$ (or more generally some prediction $\hat y = f(x_S)$ of the current instance) to compute terminal reward in \eqref{eq:reward}. 



\paragraph{Myopic discriminative heuristics (oracle guided).}
Oracle-guided methods can be interpreted as imitation learning for AFA (often via simple behavioral cloning) \citep{DBLP:conf/nips/HeDE12}. Given fully observed training samples $(x,y) \sim p(\mathbf{x},\mathbf{y})$, one simulates rollouts to generate visited states $s(x_S,y) \in \mathcal{S}$, computes an oracle action at each such state, and then fits a parametric policy to imitate the oracle’s decisions. Hence, all methods effectively come down to constructing a dataset of input/output pairs: $((S, x_S), \text{Oracle}(s(x_S,y),a))$, for many $(x,y) \sim p(\mathbf{x},\mathbf{y})$, $S \subseteq [d]$ and $a \in U$. $\text{Oracle}(s(x_S,y),a)$ represents the value of taking action $a \in U$ in state $s(x_S,y)$ according to the oracle. A model $q_{\theta}(x_S,a)$ (typically a neural net) is then trained on this dataset to predict the value of actions $a \in U$. 

This can also be understood as a form of amortized optimization \citep{10.1561/2200000102, covert2023learning}. They can also be seen as simplified variants of the RL methods discussed later: rather than learning from delayed returns, they replace credit assignment with supervised oracle labels, which makes training easier but typically biases the learned policy toward the oracle’s (often more myopic) structure. 

Finally, we assume here that the oracle signal is computed based on information in $s(x_S,y)$, so that the oracle remains fully model-free. One can also define model-based oracle signals using a model of $p(\mathbf{x}_U \mid x_S)$. However, because this requires an explicit model, we instead discuss such approaches in Section~\ref{sec:hybrid} on hybrid methods.

One of the earliest papers in this category is \citep{DBLP:conf/nips/HeDE12}. At state $s(x_S,y) \in \mathcal{S}$, they define the oracle signal (roughly) as 

\begin{equation} \label{eq:he}
\begin{aligned}
&\text{Oracle}(s(x_S,y), a) = \lambda(x_S) q_{\theta}(x_S,a) \\
&-\ell(f(x_S, x_a), y) - \alpha c_a, 
\end{aligned}
\end{equation}

where $\lambda(x_S) \in \mathbb{R}$. Episodes are simulated by rolling out the currently learned policy (i.e., via action selection based on $q_{\theta}(x_S,a)$). The coefficient $\lambda(x_S)$ is set to a large value early in an episode, when $S$ contains only a few observed features, and is gradually decreased as additional features are acquired. As with all methods discussed in this section, the predictor $f$ can be pretrained on random states $(S,x_S)$, or trained jointly with the policy.

The rationale for the first term in \eqref{eq:he} is to discourage updates that would push the learner too far from its current behavior in a single step. The paper explicitly links this idea to proximal (trust-region) methods, whose purpose is to prevent overly large policy changes during optimization \citep{10.1145/1553374.1553407, schulman2017ppo}. Moreover, their method can be viewed as a DAgger-style dataset aggregation procedure for imitation learning \citep{pmlr-v15-ross11a}. They leverage this connection to analyze the algorithm via a no-regret bound, comparing the cumulative imitation loss of the learned policy $\pi_{\theta}$ to that of the optimal policy (i.e., the policy that best imitates the oracle).

Then, after training $q_{\theta}$ on many states $s(x_S,y) \in \mathcal{S}$, it effectively learns the approximation
$q_{\theta}(x_S,a) \approx \mathbb{E}_{\mathbf{x}_a,\mathbf{y}\mid x_S}\left[-\ell\bigl(f(x_S,\mathbf{x}_a), \mathbf{y}\bigr)\right] - \alpha c_a$.
Consequently, under log loss, the policy induced by $q_{\theta}$ coincides with the myopic CMI rule in \eqref{eq:CMI}. Although \citep{DBLP:conf/nips/HeDE12} does not state this equivalence explicitly (they use a margin-based loss), subsequent work makes the connection directly, as discussed below.

\citep{gadgil2024estimating} defines the oracle signal as 

\begin{equation} \label{eq:dime}
\begin{aligned}
&\text{Oracle}(s(x_S,y), a) = \ell(f(x_S), y) -\ell(f(x_S, x_a), y).
\end{aligned}
\end{equation}

As in \citep{DBLP:conf/nips/HeDE12}, episodes are generated by rolling out the current policy, that is, by selecting actions according to $q_{\theta}(x_S,a)$. They show that, after training, $q_{\theta}(x_S,a) = I(\mathbf{y}; \mathbf{x}_a \mid x_S)$, which corresponds to the myopic CMI rule in \eqref{eq:CMI}, provided that the predictor $f$ is Bayes optimal (i.e., $f(x_S) = p(\mathbf{y} \mid x_S)$).

The works in \citep{chattopadhyay2023variational, covert2023learning, GhoshLan2023DiFA} were developed concurrently and propose very similar ideas. They define the oracle signal as $\text{Oracle}(S,x,y,a) = -\ell(f(x_S, x_a), y)$. These approaches differ from \citep{DBLP:conf/nips/HeDE12, gadgil2024estimating} in that they do not learn an action-value function $q_{\theta}$. Instead, they estimate the policy $\pi_{\theta}$ directly, in the spirit of policy gradient methods in reinforcement learning \citep{sutton2018reinforcement}. 

Concretely, they simulate episodes on training instances by rolling out $\pi_{\theta}$. At a state $(S,x_S)$, they sample an action $a \sim \pi_{\theta}(S,x_S)$ and then evaluate the resulting state by computing the loss $\ell(f(x_S, x_a), y)$. The policy parameters are then updated based on this loss. In essence, they allow gradients to backpropagate from the loss to the policy network via the reparameterization trick. This is a unique aspect of the AFA-POMDP: the terminal reward can be made differentiable with respect to the action, which enables low-variance gradient-based optimization and avoids reliance on high-variance estimators such as REINFORCE \citep{GhoshLan2023DiFA}. Moreover, \citep{covert2023learning, chattopadhyay2023variational} show that, under log loss, the resulting policy is equivalent to the selections induced by the myopic CMI criterion in \eqref{eq:CMI}. The methods differ primarily in the specific mechanism used to propagate the loss signal to the sampled action.

\paragraph{Non-myopic discriminative heuristics (oracle guided).}
The methods covered here are similar to those above, but they use a non-myopic oracle signal instead of a myopic one, yielding a non-myopic policy. The work in \citep{DBLP:journals/corr/HeMK16} extends \citep{DBLP:conf/nips/HeDE12} by defining a non-myopic oracle signal as below.

\begin{equation}
\begin{aligned}
    &\text{Oracle}(s(x_S,y), a) = -\ell(f(x_S, x_a, x_{\pi^\ast[x]}),y) \\
    &- \alpha c(\pi^\ast[x]),
    \end{aligned}
\end{equation}

where $\pi^\ast$ denotes a reference (oracle) policy and $\pi^\ast[x] \subseteq [d]$ the set of features selected by this policy on instance $x$. In other words, $q_{\theta}(x_S,a)$ is trained to predict the negative loss obtained by taking action $a \in U$ in state $(S,x_S)$ and then rolling out the remainder of the episode under $\pi^\ast$ until termination. The oracle policy is defined as the myopic heuristic in \eqref{eq:q_myop_acquire}. Consequently, although the learning objective can, in principle, support non-myopic behavior by using rollout-based targets, the resulting policy may still be biased toward myopic decisions due to the specific choice of reference policy in this paper.

\paragraph{Model-free RL.}
\citep{dulac2011datum,DBLP:journals/ml/Dulac-ArnoldDPG12} were the first to make explicit the connection between \eqref{eq:afa_obj_soft} and maximizing long-term reward in the corresponding AFA MDP (ignoring the budget constraint). Leveraging this perspective, they learn a policy in a model-free manner by simulating offline trajectories from fully observed data. The predictor and policy are trained jointly using approximate policy iteration. In their formulation, the policy $\pi_{\theta}$ takes the state $(S, x_S)$ as input, and the terminal reward is defined via 0--1 loss. 

Related model-free approaches were subsequently proposed by \citep{10.1007/978-3-642-25832-9_14,ruckstiest2013minimizing}, who learn a policy using fitted Q-iteration. Their state representation is the vector of class posterior probabilities $\{p(y_1 \mid x_S), \dots, p(y_{|\mathcal{Y}|} \mid x_S)\}$, that is, an explicit encoding of $p(\mathbf{y}\mid x_S)$, focusing on the classification setting.

More recently, this line of work has been extended to deep RL by \citep{NEURIPS2018_e5841df2,kachuee2018opportunistic,10.1609/aaai.v33i01.33013959,DBLP:journals/ml/JanischPL20}, using algorithms such as DQN \citep{mnih2015human} and PPO \citep{schulman2017ppo}. The resulting approaches are broadly similar, but differ in their design choices regarding (i) the reward signal, (ii) the state representation, and (iii) the underlying RL algorithm. Notably, while \citep{DBLP:journals/ml/JanischPL20} optimizes the objective in \eqref{eq:afa_obj_soft}, they also discuss a variant with an \emph{average} budget constraint, in which the per-instance constraint $c(\pi[x]) \leq B$ for all $x \in \mathcal{X}$ in \eqref{eq:afa_obj_soft} is replaced by $\mathbb{E}_{\mathbf{x},\mathbf{y}}\!\left[c(\pi[\mathbf{x}])\right] \leq B$. This alternative encourages allocating different costs across instances (analogous to the effect of the penalty term $\alpha\,c(\pi[\mathbf{x}])$ in \eqref{eq:afa_obj_soft}). The authors argue that, in practice, specifying a budget $B$ is often more intuitive than choosing $\alpha$.

Moreover, define $\mathbf{p}(x_S) \triangleq \{p(y_1 \mid x_S), \dots, p(y_{|\mathcal{Y}|} \mid x_S)\}$. To mitigate sparse terminal rewards, \citep{kachuee2018opportunistic} introduces an intermediate reward $R(s(x_S,y), a) = \lVert \mathbf{p}(x_S) - \mathbf{p}(x_S, x_a) \rVert / c_a$ for $a \in U$, which quantifies model change via the shift in the predictive distribution $p(\mathbf{y}\mid x_S)$, normalized by the feature cost $c_a$. While this shaping can bias learning toward more myopic behavior, it may improve training stability. In addition, \citep{NEURIPS2018_e5841df2,kachuee2018opportunistic} propose sharing parameters between the predictor $f$ and the policy $\pi$.


\subsection{Hybrid Methods} \label{sec:hybrid}

The methods in this section are fundamentally model-free, similar to the methods in Section \ref{sec:model_free}. However, they also utilize a model (based on $p(\mathbf{x}_U \mid x_S)$) during training, typically for the purpose of making training more stable.

\paragraph{Myopic hybrid heuristics.}

All methods in the class of \emph{myopic generative heuristics} (Section~\ref{sec:model_based_methods}) can be used to define an oracle signal. For instance, \citep{ma2019eddi} estimate the conditional mutual information $I_{\omega}(\mathbf{y}; \mathbf{x}_a \mid x_S)$ by explicitly modeling $p(\mathbf{x}_U \mid x_S)$ with a deep arbitrary conditional generative model, parameterized by $\omega$. This naturally yields the oracle:

\begin{equation}
\text{DeployableOracle}(b(x_S), a) = I_{\omega}(\mathbf{y}; \mathbf{x}_a \mid x_S).
\end{equation}
The advantage is that, once $q_{\theta}$ has learned to predict this quantity, information gain can be approximated for test instances using simple forward passes through $q_{\theta}$, rather than computing $I_{\omega}(\mathbf{y}; \mathbf{x}_a \mid x_S)$ directly at inference time, which typically involves costly Monte Carlo estimation. This can substantially reduce test-time acquisition cost, which is valuable in applications where computational efficiency is critical. 

Moreover, a unique property of such oracles is that they are \emph{deployable}, meaning the oracle signal only depends on the observed part of the state (e.g., components of the belief state $b(x_S)$ in \eqref{eq:beliefstate}). This distinction was originally made in \citep{DBLP:conf/icml/ValanciusLO24}. While their oracle signal is non-myopic (and thus covered in the next subsection), they motivate deployable oracles on the grounds that (i) the oracle itself can be executed and evaluated at inference time (unlike ``cheating'' oracles that condition on unobserved parts of the state $s(x_S,y)$ (i.e., $x_U$ and $y$), which allows one to measure oracle quality directly, and (ii) when the oracle is used as a teacher for imitation learning, the teacher and student share the same inputs, helping disentangle imitation error from oracle suboptimality and typically yielding a more reliable learned acquisition policy. 

However, an oracle in this category does not have to be deployable since it could also depend on $(x_U, y)$ (in addition to explicit dependence on a model $p(\mathbf{x}_U \mid x_S)$).

There is no paper that explicitly fits into this category, but is a possible direction for future research. 

\paragraph{Non-myopic hybrid heuristics.} In Section ~\ref{sec:model_based_methods} under \emph{non-myopic generative heuristics}, we introduced two methods by \citep{DBLP:conf/icml/ValanciusLO24, huang2025informationtemplatesnewparadigm}. However, the utilities computed by these methods can also serve as an oracle signal that $q_{\theta}$ learns to predict directly. Notably, their oracle is \emph{deployable} (see discussion above), since computing $\Delta_k^{\text{NA}}(a, x_S)$ in \eqref{eq:nonadaptive} only depends on observable parts of the state (e.g., components of the belief state $b(x_S)$).
 
\citep{guney2025active} first trains a predictor $f$ on a fully observed training set, without feature masking. Given the trained predictor $f$, it then computes a local explanation vector for each instance, $\phi(x) \in \mathbb{R}^d$ (for example, SHAP or LIME feature attributions), where $\phi(x,a)$ denotes the importance assigned to feature $a$ for instance $x$. The oracle signal in state $s(x_S,y)$ for taking action $a \in U$ is defined as
\begin{equation} \label{eq:shaporacle}
    \text{Oracle}(s(x_S,y), a) = \phi(x,a),
\end{equation}
so that $q_\theta$ is trained to predict local explanation scores for test instances.

Producing $\phi(x)$ for a test instance $x \in \mathcal{X}$ requires access to all features in $x$, which is not available under partial observability. Hence, the oracle in \eqref{eq:shaporacle} is not deployable in the sense of \citep{DBLP:conf/icml/ValanciusLO24, huang2025informationtemplatesnewparadigm}. Moreover, the method used to generate $\phi(x)$ given $f$ implicitly models $p(\mathbf{x}_U \mid x_S)$ (to capture feature interaction) and is hence not a fully model-free oracle. Finally, because some explanation methods (notably SHAP) can reflect feature interactions and joint informativeness, the resulting signal may be non-myopic to a limited extent.

Finally, \citep{GhoshLan2023DiFA} was presented as a \emph{myopic discriminative heuristic} in Section~\ref{sec:model_free}. However, the authors also describe an optional extension that augments the policy state representation (the input to $\pi_{\theta}$) with imputations from a generative model $p(\mathbf{x}_U \mid x_S)$. This variant is therefore model-based and could, in principle, support non-myopic behavior, although the paper does not explicitly analyze this aspect.

\paragraph{Model-based RL.}
Some RL-based methods learn a model of the conditional distribution $p(\mathbf{x}_U \mid x_S)$. This model is primarily used to improve training stability. For example, \citep{zannone2019odin} generates synthetic trajectories by sampling from $p(\mathbf{x}_U \mid x_S)$, using the PVAE model of \citep{ma2019eddi}. In contrast, \citep{pmlr-v139-li21p,li2024distributionguidedactivefeature} leverage the learned distribution by enriching the policy input with imputations or summary statistics of $p(\mathbf{x}_U,\mathbf{y} \mid x_S)$ and by providing denser reward signals. While such shaping can bias the learned policy toward more myopic behavior, it often yields more stable optimization. They model $p(\mathbf{x}_U,\mathbf{y} \mid x_S)$ using the ACFlow model proposed in \citep{pmlr-v119-li20a}.

\citep{9533593} first learns a policy in a model-free setting using an advantage actor-critic algorithm, and then refines it with Monte Carlo tree search \citep{browne2012mcts}. The learned policy is used as the rollout policy and, in effect, provides a model for the transition dynamics during search.

Although these approaches estimate aspects of the dynamics, in the sense of learning $p(\mathbf{x}_U \mid x_S)$, their core training procedure remains model-free RL. The learned model is mainly used as an auxiliary component to stabilize learning, making these methods hybrid between model-free and model-based paradigms.

\section{Conclusion and Future Directions}
\label{sec:conclusion}

AFA studies how to sequentially acquire feature values under costs and decide when to stop and predict. In this survey, we organize modern AFA around its underlying POMDP structure, showing that most methods can be understood as different approximations to the same sequential decision problem. We introduced a unified taxonomy that separates (i) embedded cost-aware predictors, (ii) model-based methods, (iii) model-free methods, and (iv) hybrid methods. We hope this POMDP-centric view both clarifies existing work and motivates new AFA methods that more directly build on the mature literature on POMDP planning and approximation. In particular, online planning is an effective way to handle large, expensive POMDPs \citep{10.5555/1622673.1622690}, and is underutilized in AFA. Despite substantial progress, several research directions remain open, which we now discuss.

\paragraph{AFA Benchmarking.} Recent myopic discriminative heuristics \citep{covert2023learning, gadgil2024estimating} often perform remarkably well on common AFA datasets, in many cases outperforming non-myopic approaches, including RL-based methods \citep{schütz2025afabenchgenericframeworkbenchmarking}. At the same time, the AFA benchmark in \citep{schütz2025afabenchgenericframeworkbenchmarking} shows that when datasets are synthetically constructed so that optimal performance requires non-myopic selection, non-myopic methods expectedly outperform myopic ones. This motivates a more principled understanding of when non-myopic policies are necessary for a given data distribution \(p(\mathbf{x}, \mathbf{y})\), and whether such insights can be used to decide when the additional complexity of non-myopic planning is justified. In addition, empirical studies of modern AFA methods rarely include the cost-sensitive decision trees described in Section~\ref{sec:embedded} as baselines. These models can be viewed as simple policies for AFA, and including them helps assess whether more advanced approaches offer meaningful improvements beyond such straightforward alternatives.

\paragraph{Connection to decision trees.} In Section~\ref{sec:embedded}, we explained that any deterministic AFA policy $\pi$ can be represented as a decision tree. Hence, the vast literature on decision trees \citep{Costa2023RecentAdvancesDecisionTrees} address a closely related problem to AFA. The key difference is that standard decision tree learning typically does not account for feature acquisition costs. While most classical decision tree algorithms are myopic, recent work has increasingly focused on constructing non-myopic trees (that is, \emph{optimal decision trees}), enabled by improved computational resources compared to the early era of decision tree methods \citep{Costa2023RecentAdvancesDecisionTrees}. Given this conceptual overlap, advances in (optimal) decision tree learning may inform future AFA methods, and insights from AFA can likewise motivate new directions in decision tree research.

\paragraph{Connection to local explanability methods.} A wide range of methods have been proposed for assigning local feature-importance scores to individual instances $x \in \mathcal{X}$ for interpretability purposes, see \citep{molnar2025}. In addition, as discussed in Section~\ref{sec:methods}, an AFA policy $\pi$ can itself serve as an instance-level explanation by indicating which features $\pi[x]$ are acquired for a given $x$. Several AFA works therefore incorporate interpretability as an explicit design goal, referring to predictors $f$ equipped with a corresponding AFA policy $\pi$ as \emph{interpretable by design} \citep{chattopadhyay2022interpretable,chattopadhyay2023variational}. Related to this, \citep{li2024distributionguidedactivefeature} proposes goal-based explainable AFA, where they modify AFA so that each acquisition step can be explained as pursuing an explicit, human-readable sub-goal, rather than only improving the prediction.

\paragraph{Types of feature cost.} The cost of acquiring feature \( a \in [d] \) is \( c_a \). In many works on AFA discussed above, it is common to assume \emph{uniform} feature costs, i.e., \( c_a = 1 \) for all \( a \in [d] \). However, in practice, feature costs are often \emph{non-uniform}, meaning \( c_a \neq c_b \) for some \( a, b \in [d] \). Certain methods can naturally incorporate such heterogeneous costs, while others cannot. Recently, stochastic feature costs have also been explored in the AFA setting~\citep{rahbar2025costefficient}. 

\paragraph{Misclassification cost.} Beyond feature acquisition costs, cost-sensitive learning can incorporate many other notions of cost \citep{turney2002typescostinductiveconcept}. A common example is non-uniform \emph{misclassification cost}, considered by several methods discussed in Section~\ref{sec:methods} (e.g., \citep{DBLP:journals/jair/ZubekD05,10.5555/1622826.1622838,DBLP:conf/aaai/BilgicG07,9414669}). The problem formulation of AFA used in this survey subsumes this setting, since misclassification costs can be incorporated directly into the loss function $\ell$. In this setting, different types of errors incur different penalties, reflecting that misclassifying certain classes can be more consequential than misclassifying others.

\paragraph{Generative modeling for partially observed inputs.} As discussed in Section~\ref{sec:model_based_methods}, learning a model of $p(\mathbf{x}_U \mid x_S)$ can strengthen AFA methods. Consequently, progress on estimating $p(\mathbf{x}_U \mid x_S)$, whether via deep arbitrary conditional generative models or simpler approaches, can directly benefit the development of model-based AFA.

\paragraph{Realistic evaluation.} Most AFA papers assume an offline setting with fully observed feature vectors, split into training and test sets for learning and evaluation. This can conflict with the premise that features are costly, and real datasets often have missing values (e.g., medical records include only ordered tests). Consequently, if a policy acquires a feature that is systematically missing in the logged data, it cannot be reliably evaluated. This motivates evaluation protocols that account for realistic logging and missingness. \citep{vonkleist2023evaluationactivefeatureacquisition} studies this issue in the standard AFA setting considered in this survey. \citep{JMLR:v26:23-1635} investigates related challenges for time series data, where the value of a feature $\mathbf{x}_a$ may change over time, making the timing of measurements essential (discussed further below).

\paragraph{Permutation-invariant deep set encoders.} Several AFA methods discussed in Section~\ref{sec:methods} (e.g., \citep{ma2019eddi,NEURIPS2018_e5841df2}) represent the current state $(S, x_S)$, used as input to both the predictor and the policy, using permutation-invariant deep set encoders \citep{Vinyals2015OrderMatters,8099499}. These architectures map sets of varying cardinality to fixed-dimensional latent representations. This is particularly relevant for AFA, where the input is a partially observed feature vector whose observed dimension varies with $S$. A promising research direction is therefore to study, in a systematic way, when such set encoders provide measurable benefits over simpler masking-based representations (Section~\ref{sec:problem_formulation}).

\paragraph{Performance guarantees.} Third, most AFA methods come without performance guarantees. A notable exception constitutes the methods based on the EC$^2$ objective \citep{ijcai2023p463, rahbar2025costefficient}, which admit near-optimality guarantees, but the objective is typically expensive to compute. See further discussion on performance guarantees in Section \ref{sec:myopic_nonmyopic}.

\paragraph{AFA for time series data (longitudal AFA).} A growing body of work studies AFA in time series settings \citep{yoon2018deep,pmlr-v106-yoon19a,yin2020reinforcementlearningefficientactive,pmlr-v119-jarrett20a,qin2023riskaverse,dinh2025nocta,JMLR:v26:23-1635}. Here, an instance (for example, a patient) is not described by a single static feature vector $x \in \mathcal{X}$, but by a time-indexed trajectory $x(t) \in \mathcal{X}$ over $t \in \{1,\dots,L\}$. At each time step $t$, the policy acquires one or more features from $x(t)$ based on the acquisition history up to time $t-1$, that is, observations from $x(1), \dots, x(t-1)$. 

This setting differs from standard AFA in several important ways. First, features may be acquired repeatedly, since their values can evolve over time (for example, a patient's blood pressure depends on when it is measured). More broadly, the underlying state of the instance changes over time and may also be influenced by earlier measurements or interventions. Second, the objective often extends beyond cost-efficient accuracy to include \emph{timeliness}, that is, making accurate predictions as early as possible. In medical applications, for instance, the goal may be to reach a reliable diagnosis quickly rather than only at the end of the observation window. Third, time series formulations can capture delayed measurements, where a feature requested at time $t$ is only observed at time $t+v$, with $v$ denoting the delay.

\paragraph{Stopping criteria.} The POMDP formulation of AFA provides a principled mechanism for deciding when to stop acquiring features and output a prediction (Sections~\ref{sec:pomdp_formulation} and \ref{sec:myopic_nonmyopic}). In practice, however, the AFA-POMDP is intractable, and many works therefore rely on heuristic acquisition rules that are often myopic (e.g., \eqref{eq:q_myop_acquire}). When acquisition is highly myopic, as in \eqref{eq:q_myop_acquire}, it is less clear that the POMDP-derived stopping criterion remains appropriate, and alternative stopping rules have been explored. A common choice is to stop once the entropy of $p(\mathbf{y} \mid x_S)$ falls below a threshold. Another is to stop when the estimated value of every remaining unobserved feature (for example, $I(\mathbf{y}; \mathbf{x}_a \mid x_S)$) is below a threshold \citep{gadgil2024estimating}. As a result, many AFA methods use fixed or learned thresholds to govern the stopping decision. This stopping problem is closely related to the classical theory of \emph{optimal stopping} \citep{ChowRobbinsSiegmund1971}, which studies when to take an action so as to maximize expected reward or minimize expected cost.

\paragraph{Batch selection.} Some studies on AFA consider the setting of \emph{batch selection}, where the acquisition policy \( \pi \) selects a batch of \( k \) features at each step rather than a single feature \citep{DBLP:conf/icml/ShengL06, pmlr-v28-chen13b, 10.1609/aaai.v38i10.28967}. This concept is also well studied in the related field of \emph{active learning} \citep{10.1145/3472291}. The main motivation for batch selection in AFA is practical. In the non-batch case (\( k = 1 \)), it is typically assumed that the true feature value becomes available immediately after acquisition. However, in many real-world applications there may be a substantial delay between querying a feature and observing its value. In such cases, acquiring multiple features in parallel can be more efficient. For instance, in healthcare settings, medical tests often require significant processing time, so obtaining the results of several tests simultaneously may enable faster decisions. This can be crucial in many medical or other time-sensitive settings.  

Each selection step can be viewed as \emph{non-adaptive} selection of $k$ features conditioned on the features already acquired. This means that at each step, a set of \( k \) features is selected non-adaptively (essentially performing static feature selection), while the overall process remains adaptive to the outcomes of previously acquired features. As a result, the resulting policy typically performs worse than a fully adaptive approach (\( k = 1 \)), since it cannot exploit intermediate feedback to refine subsequent feature selections.

\paragraph{AFA for unstructured and high-dimensional data.} The complexity of AFA grows exponentially with the number of features $d$, making naive per-feature acquisition impractical in high-dimensional settings. A common remedy is to impose structure by grouping features and acquiring groups rather than individual features. This is closely related to batch selection, but with a fixed set of allowable batches determined by the grouping \citep{li2024distributionguidedactivefeature,takahashi2025dynamicfeatureselectionvariable}. This approach has been used for AFA on image data, for example in \citep{GhoshLan2023DiFA,gadgil2024estimating,guney2025active}. There, an image is partitioned into patches, and the policy acquires patches sequentially instead of querying individual pixels.

\paragraph{AFA for regression.} Most AFA work focuses on classification, even though the formulation adopted in this survey applies equally to regression. A few of the methods discussed above address both settings (e.g., \citep{pmlr-v139-li21p}), but regression remains comparatively underexplored. Developing and systematically evaluating AFA methods for regression therefore represents a promising direction for future research.

\paragraph{Cascades and trees of classifiers.}
A related line of work studies \emph{cascades} and \emph{trees} of classifiers for cost-aware inference \citep{ViolaJones2004RobustRealTimeFaceDetection,2013cstc,xu14a,NIPS2017_d9ff90f4}. Both can be seen as a tree whose nodes are classifiers that use a designated subset of features, where going deeper typically means using additional (often more expensive) features. A \emph{cascade} is a single chain: at each stage one either exits or moves to the next stage, so there is effectively only one continuation branch. Cascades are therefore most closely related to AFA variants that \emph{fix the acquisition order} and only decide when to stop \citep{9422113}. A \emph{tree} allows multiple branches and is thus conceptually closer to adaptive AFA, but the branching structure is usually fixed (often a shallow binary tree) and learning optimizes predictors, feature subsets, and gating rules \emph{within} that structure \citep{2013cstc,xu14a,NIPS2017_d9ff90f4}. In contrast, AFA learns a general acquisition policy over partial observations without committing to a predetermined routing topology.

\paragraph{Choice of loss function $\ell$.} The most common loss function used in the reward \eqref{eq:reward} is the log loss by far, resulting in $r(b(x_S), \texttt{STOP}) = -H(\mathbf{y} \mid x_S)$ (see \eqref{eq:reward}). However, it is not clear that reducing entropy is always the best choice for AFA \citep{norcliffe2025stochasticencodingsactivefeature}. To see this, consider two three-class distributions: \(H([0.5,0.5,0])=\ln 2\approx 0.693\) and \(H([0.7,0.15,0.15])\approx 0.819\). The first has lower entropy yet leaves a tie between two classes, while the second identifies a unique top class. Entropy can be decreased by driving some class probabilities toward zero without sharpening the distinction among the leading labels. When the goal is to identify the most likely class, the acquisition objective should focus on separating the top candidates. A solution is to consider objectives that put more weight on the most likely class, such as the 0--1 loss objective \(r(b(x_S), \texttt{STOP})=\max_y p(y\mid x_S)\). We can also replace Shannon entropy with the R\'enyi entropy \citep{renyi1961_measures_entropy_information}
\[
H_\alpha(p)=\frac{1}{1-\alpha}\log \sum_y p(y\mid x_S)^\alpha,
\]
which recovers Shannon entropy as \(\alpha\to 1\) and converges to the min-entropy \(H_\infty(p)=-\log \max_y p(y\mid x_S)\) as \(\alpha\to\infty\), thereby emphasizing the most likely class.

\paragraph{Other special settings of AFA.} Finally, several specialized AFA variants were not discussed in depth above, which we briefly note here. These include: (i) meta AFA, where a policy $\pi$ is trained across many tasks and then adapted to a target task via in-context learning \citep{kobayashi2025learningtomeasureincontextactivefeature}; (ii) causal AFA, where the objective shifts from accurate prediction to selecting treatments or interventions, often in clinical settings and sometimes studied under the names \emph{dynamic testing regimes} or \emph{dynamic treatment regimes} \citep{Neugebauer2017JointEffect,Liu02012021,pmlr-v258-piskorz25a}; (iii) two-stage acquisition schemes that additionally query costly expert decisions and treat expert outputs as observations in the belief update \citep{10447423}; (iv) AFA with external classifiers \citep{10097057}; (v) AFA under a fixed acquisition order \citep{9422113}; and (vi) AFA with out-of-distribution detection \citep{li2024distributionguidedactivefeature}.

\section*{Acknowledgments}

This work was partially supported by the Wallenberg AI, Autonomous Systems and Software Program (WASP) funded by the Knut and Alice Wallenberg Foundation. We would like to thank Daphney–Stavroula Zois for pointing us to a number of relevant works.

\bibliographystyle{named}
\bibliography{ijcai26}
\end{document}